\documentclass[10pt,journal,compsoc]{IEEEtran}
\usepackage{soul,framed} %,caption
\usepackage{colortbl}
\usepackage[dvipsnames]{xcolor}
\usepackage[pdftex]{graphicx}
\graphicspath{{../pdf/}{../jpeg/}}
\DeclareGraphicsExtensions{.pdf,.jpeg,.png}

\usepackage[cmex10]{amsmath}
\usepackage{array}
\usepackage{mdwmath}
\usepackage{mdwtab}
\usepackage{eqparbox}
\usepackage{url}
\usepackage{graphicx}
\usepackage{amsmath}
\usepackage{amsfonts}
\usepackage{algorithm}
\usepackage{algpseudocode}
\usepackage{multirow}
\usepackage{footnote}
\usepackage{bm}
\usepackage{makecell}
\usepackage{pdflscape}
\usepackage{afterpage}
\usepackage{bbding}
\usepackage{lscape}
\usepackage{booktabs}
\usepackage{tikz}
\usepackage{ragged2e}
\usepackage{amssymb}% http://ctan.org/pkg/amssymb
\usepackage{pifont}% http://ctan.org/pkg/pifont
\newcommand{\xmark}{\ding{55}}%

\usepackage{pgfplots}
\usepackage{subcaption}

\usepackage[square,sort,comma,numbers]{natbib} % multiple citation or 
\usepackage[top=0.4in,bottom=0.4in,left=0.4in,textwidth=7.7in]{geometry}

\usepackage[pagebackref=false,breaklinks=true,linkcolor=red,anchorcolor=black, citecolor=green,colorlinks,bookmarks=true]{hyperref}
\usepackage[capitalize]{cleveref}
\crefname{section}{Sec.}{Secs.}
\Crefname{section}{Section}{Sections}
\Crefname{table}{Table}{Tables}
\crefname{table}{Tab.}{Tabs.}

\newcommand{\etal}{\textit{et al}. }
\newcommand{\etc}{\textit{etc}. }
\newcommand{\ie}{\textit{i}.\textit{e}., }
\newcommand{\eg}{\textit{e}.\textit{g}., }
\usepackage{amsfonts}
\usepackage{pifont}

\pgfplotsset{
    layers/my layer set/.define layer set={
        background,
        main,
        foreground
    }{ },
    set layers=my layer set,
}
\pgfplotsset{compat=1.12}

\begin{document}

\title{Lightweight Pixel Difference Networks for Efficient Visual Representation Learning}
 
\author{\IEEEauthorblockN{
Zhuo Su, Jiehua Zhang\thanks{
Zhuo Su is with the College of Computer Science, Nankai University, Tianjin
300071, China, and also with the Center for Machine Vision and Signal Analysis
(CMVS), University of Oulu, 90570 Oulu, Finland (e-mail: zuike2013@outlook.com). \\
Jiehua Zhang is with the Center for Machine Vision and Signal
Analysis (CMVS), University of Oulu, 90570 Oulu, Finland (e-mail:
jiehua.zhang@oulu.fi).\\
Longguang Wang is with the Aviation University of Air Force, Changchun,
Jilin 130012, China (e-mail: wanglongguang15@nudt.edu.cn).\\
Hua Zhang is with the Institute of Information Engineering, Chinese Academy
of Sciences, Beijing 100045, China (e-mail: zhanghua@iie.ac.cn).\\
Zhen Liu and Li Liu are with the College of Electronic Science and Technology, Nation University of Defense Technology, Changsha, Hunan 410000,
China (e-mail: zhen\_liu@nudt.edu.cn).\\
Matti Pietikäinen is with the Center for Machine Vision and Signal Analysis,
University of Oulu, 90014 Oulu, Finland (e-mail: matti.pietikainen@oulu.fi).\\
\\
The code is available at  \href{https://github.com/hellozhuo/pidinet}{https://github.com/hellozhuo/pidinet}.\\
Corresponding author: Li Liu (dreamliu2010@gmail.com)},
Longguang Wang,
Hua Zhang,
Zhen Liu,
Matti Pietik\"{a}inen, 
Li Liu 
}}

% The paper headers
% \markboth{ In preparation for submitting to IEEE TPAMI}%
%\markboth{Revision Submitted to IEEE TPAMI}
%\markboth{IEEE TRANSACTIONS ON PATTERN ANALYSIS AND MACHINE INTELLIGENCE}
\markboth{IEEE TRANSACTIONS ON PATTERN ANALYSIS AND MACHINE INTELLIGENCE, DOI: 10.1109/TPAMI.2023.3300513}
{Su \MakeLowercase{\textit{et al.}}: }

\IEEEtitleabstractindextext{%
\begin{abstract}
\justifying
% Li's compact abstract
Recently, there have been tremendous efforts in developing lightweight Deep Neural Networks (DNNs) with satisfactory accuracy, which can enable the ubiquitous deployment of DNNs in edge devices. The core challenge of developing compact and efficient DNNs lies in how to balance the competing goals of achieving high accuracy and high efficiency. In this paper we propose two novel types of convolutions, dubbed \emph{Pixel Difference Convolution (PDC) and Binary PDC (Bi-PDC)} which enjoy the following benefits: capturing higher-order local differential information, computationally efficient, and able to be integrated with existing DNNs. With PDC and Bi-PDC, we further present two lightweight deep networks named \emph{Pixel Difference Networks (PiDiNet)} and \emph{Binary PiDiNet (Bi-PiDiNet)} respectively to learn highly efficient yet more accurate representations for visual tasks including edge detection and object recognition. Extensive experiments on popular datasets (BSDS500, ImageNet, LFW, YTF, \emph{etc.}) show that PiDiNet and Bi-PiDiNet achieve the best accuracy-efficiency trade-off.  For edge detection, PiDiNet is the first network that can be trained without ImageNet, and can achieve the human-level performance on BSDS500 at 100 FPS and with $<$1M parameters. For object recognition, among existing Binary DNNs, Bi-PiDiNet achieves the best accuracy and a nearly $2\times$ reduction of computational cost on ResNet18.
\end{abstract}

}

% make the title area
\maketitle
\IEEEdisplaynontitleabstractindextext
\IEEEpeerreviewmaketitle

%%%%%%%%% BODY TEXT
\IEEEraisesectionheading{
\section{Introduction}
\label{sec:intro}}

% *****Significance*****

% From Li
\textcolor{black}{
During the past decade, DNNs, especially deep convolutional neural networks (DCNNs), have revolutionized many computer vision tasks including edge detection~\cite{xie2017holistically,liu2019richer,he2019bidirectional}, image segmentation~\cite{minaee2021segmentationsurvey}, object recognition~\cite{liu2017sphereface,he2016residual}, and object detection~\cite{liu2020deep}. Much of this progress has been enabled by increasingly large
and energy hungry DNNs \cite{krizhevsky2017alexnet,simonyan2014vggnet,he2016residual,huang2019gpipe,DBLP:conf/iclr/vit,zoph2018learning}. 
Despite high accuracy, computationally expensive DNNs cause serious issues for  sustainability, environmental friendliness, broad economic viability, and their ubiquitous deployment on edge devices like drones and embedded/wearable/IoT devices that have very limited computing resources and low power. Therefore, there has been a wide range of interest in developing techniques via algorithm and hardware optimization in order to enable efficient implementation of DNNs for improved energy efficiency. Numerous methods for DNN compression and acceleration  have been proposed, and the mainstreams include compact network design \cite{howard2017mobilenets,zhang2018shufflenet, zhang2022parcnet, mehta2021mobilevit}, model quantization~\cite{zhou2016dorefa,lin2017abcnet,liu2021post, jeon2022mr},
tensor decomposition~\cite{ouerfelli2022random,pmlr-v162-wang22ar}, network pruning~\cite{han2015deepcompression,he2017channelpruning,li2022revisiting}, 
knowledge distillation~\cite{touvron2021deit, gou2021knowledge}, and efficient neural architecture search~\cite{ding2021hr,lee2021hardware}.
}

\textcolor{black}{Among the aforementioned mainstreams, compact model design and model quantization have received significant attention. Compact model design aims at directly creating a more efficient network architecture with significantly reduced computational cost while maintaining accuracy as much as possible, such as MobileNet \cite{howard2017mobilenets}, ShuffleNet \cite{zhang2018shufflenet}, MobileViT \cite{mehta2021mobilevit}, and ParCNet \cite{zhang2022parcnet}, and enjoys the benefits of achieving both training and inference efficiency. For a given task, compact models can be designed by various techniques  such as more complex but compact
branch topology~\cite{szegedy2016inceptionv2,cheng2021csnet} and more flexible convolution operators like depthwise separable convolution~\cite{chollet2017xception,sandler2018mobilenetv2} and grouped convolution~\cite{xie2017aggregated,su2020dynamic}. However, existing compact models use conventional convolutions that have limited expressive power (which will be discussed later), restricting the diversity of extracted feature maps and leads to suboptimal efficiency-accuracy balance.}
\textcolor{black}{Model quantization~\cite{zhou2016dorefa,lin2017abcnet,liu2021post} aims for inference efficiency and attempts to shrink the size of a given DNN model for saving memory storage and computation by data quantization, \emph{i.e.}, reducing the bitwidth of the model weights and activations. As an ambitious case of model quantization, the Binary Convolutional Neural Networks (BCNNs)~\cite{courbariaux2016bnn,zhou2016dorefa,liu2020reactnet} binarize model weights and activations, which result in significant saving of computing resources. However, BCNNs suffer from big accuracy drops in comparison with their real-valued versions. Therefore, how to develop highly accurate BCNNs remains open.}

\begin{figure}[t!]
    \centering
    \includegraphics[width=0.96\linewidth]{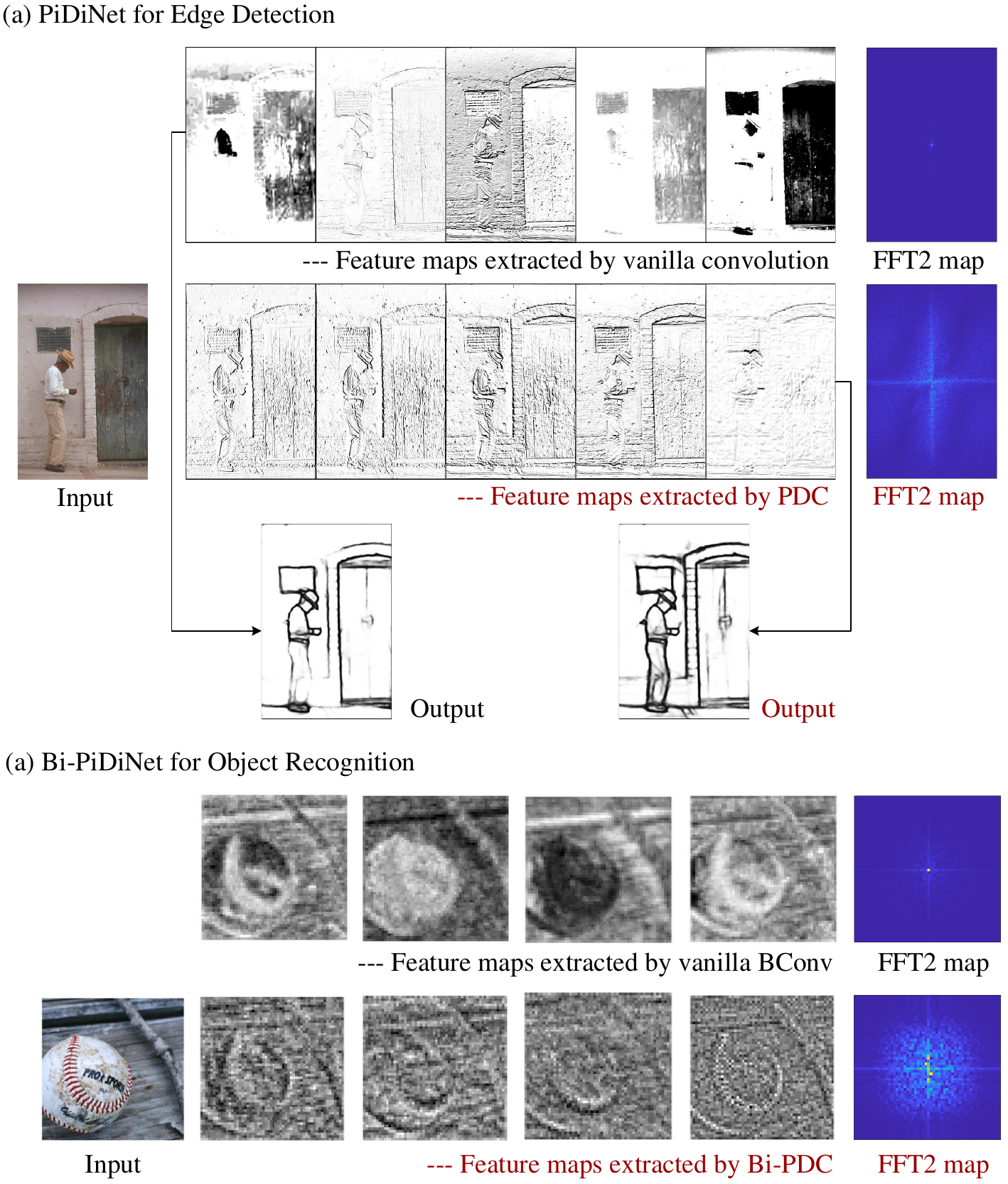}
    \caption{Compared with vanilla (binary) convolution, the proposed (binary) PDC can better capture high-order local differential information that facilitates edge detection and general object recognition. In frequency domain, the high-order information contains more high-frequency components, as illustrated in the FFT2 map averaged over all the feature maps.}
    \label{fig:figure2}
\end{figure}

\textcolor{black}{To advance the state of the art in efficient DCNNs, we effectively look
inside the internal structure of deep features. The expressive power of a DCNN is related to two main structures: network
depth and convolution~\cite{krizhevsky2017alexnet,szegedy2015inceptionv1}. 
For the former, a DCNN learns a sequence of hierarchical representations that correspond to increasing abstraction levels.  The latter is convolution, probing image patterns via the use of translation invariant, local operators, which corresponds to the extraction of local descriptors in traditional shallow image representation frameworks like Bags of Words (BoW)~\cite{liu2019bow}. It has been generally recognized that various local descriptors like Local Binary Patterns (LBP)~\cite{ojala2002multiresolution,liu2012extended},  Histogram of Oriented Gradients (HOG)~\cite{william1994hog}, Sorted Random Projections (SRPs)~\cite{liu2011sorted} are robust and discriminative for describing fine-grained image information, but their role may be limited by  the traditional shallow BoW pipeline. However, by contrast, the conventional convolution underlying DCNNs  only captures the pixel intensity cues while failing to  encode image microstructure like higher-order local gradient information (as shown in \cref{fig:figure2}). } 

Thereby, a natural question arises: why not combine the best of both worlds, namely integrating traditional local descriptors into DCNNs? To our best knowledge, this direction has not received enough attention and is worth future exploration, as we realized that such higher-order local differential 
information, which is ignored by conventional convolution, can well capture microtexture information and has been proven to be powerful before deep learning. For instance, the gradient information has been demonstrated to be effective in numerous edge detectors~\cite{sobel19683x3,prewitt1970object,gupta2014seng}. 
The higher-order microtexture cues have been shown to be highly successful in conventional facial and texture recognition methods like LBP methods~\cite{ojala2002multiresolution,liu2012extended,su2019bird,7393828mrelbp,liu2011sorted}.

% ***** Our solution/method *****
Based on the  aforementioned motivations and our previous work in (Extended) LBP~\cite{ojala2002multiresolution,liu2012extended,su2019bird} and SRPs~\cite{liu2011sorted}, 
\textcolor{black}{
in this work, we aim to develop a generic convolution operation, namely, PDC, to benefit the widely used CNN architectures for vision applications. Our PDC is designed by integrating the LBP mechanism into the basic convolution operations to enable the filters to probe local pixel differences rather than pixel intensities. With different LBP probing strategies, we develop three PDC instances, namely, Central PDC, Angular PDC, and Radial PDC to capture rich higher-order feature statistics from different encoding directions. 
}

\textcolor{black}{
Generally, PDC has three remarkable properties. \textbf{First}, it is capable of producing features with high-order information that are complementary to those obtained by vanilla convolutions, which enriches the diversity of feature maps (\cref{fig:figure2}).  \textbf{Second}, it is fully differentiable and can be seamlessly incorporated into any network architecture for end-to-end optimization. \textbf{Third}, it is compatible with existing network acceleration techniques like network binarization to obtain further efficiency gains (Bi-PDC in \cref{sec:Bi-PDC}).
}

\textcolor{black}{
To demonstrate the effectiveness of our method in both semantically low-level tasks (\ie edge detection) and high-level tasks (\ie image classification and facial recognition), we build two highly efficient DCNN architectures, where PDC and Bi-PDC are used. Extensive experiments show that our PDC facilitates CNNs to achieve a more desirable balance between efficiency and accuracy.
}

\textcolor{black}{This paper is a substantial extension of our published conference paper \cite{su2021pidinet}\footnote{Our preliminary version was  published in ICCV as an oral paper.}. The main contributions of this work are summarized as follows.}
\begin{itemize}
    \item \textcolor{black}{We propose two types of convolutions, PDC and Bi-PDC, that can capture the  higher-order local differential information complementary to that by vanilla convolution. They are computationally efficient and can be integrated well into existing DCNNs.}
    \item \textcolor{black}{With the proposed PDC, we design a novel compact DCNN architecture, dubbed Pixel Difference Network (PiDiNet), for the edge detection task. To our best knowledge, PiDiNet is the first deep network that can achieve the human-level performance without ImageNet pretraining (on the popular BSDS500 dataset).} 
    \item \textcolor{black}{With the proposed Bi-PDC,  we further propose a new binary architecture named Binary Pixel Difference Networks (Bi-PiDiNet)  that can flexibly fuse Bi-PDC with vanilla binary convolution to capture both the zeroth-order and higher-order local image information for efficient object recognition. Bi-PiDiNet is designed to be more compact yet more accurate.} 
    
    \item \textcolor{black}{Extensive experimental evaluations on the commonly used datasets for edge detection, image classification, and facial recognition demonstrate that our proposed PiDiNet and Bi-PiDiNet achieve higher accuracy yet improved efficiency compared with the state of the art. The proposed PiDiNet and Bi-PiDiNet advance the potential of using highly lightweight deep models for realizing efficient  vision tasks at the edge.}
\end{itemize}

The rest of the paper is organized as follows. In Section~\ref{sec:related_work}, we review related works. In Section~\ref{sec:pdc} and Section~\ref{sec:Bi-PDC}, we elaborate on the proposed PDC and Bi-PDC in great detail, respectively. A theoretical interpretation of PDC and Bi-PDC is given in Section~\ref{sec:theoretical}. In Section~\ref{sec:application}, we develop two task-specific architectures based on the proposed convolutions. Following that, the experimental results are presented in Section~\ref{sec:experiments}. Finally, we conclude our paper in Section~\ref{sec:conclusion}.

\section{Related Work}
\label{sec:related_work}

\subsection{Efficient Visual Representation Learning}
Great efforts have been taken in the last decades to achieve efficient visual representation learning, which can be categorized into the following aspects.

\vspace{0.3em}
\noindent \textbf{Compact model design.} To build models with less computational cost and memory storage, many lightweight CNNs have been developed~\cite{howard2017mobilenets,zhang2018shufflenet,tan2019efficientnet,mehta2021mobilevit,zhang2022parcnet}. 
These networks were directly built from scratch by using compact modules like depthwise separable convolution~\cite{howard2017mobilenets,chollet2017xception}, grouped convolution~\cite{xie2017aggregated,su2020dynamic}, deformable convolution~\cite{zhu2019deformable}, and compact vision attention modules~\cite{mehta2021mobilevit}, which are memory and computation friendly via sparse structures. For example, a depthwise separable convolutional layer conducts spatial and cross-channel correlations individually to save the overall computation. Grouped convolution divides the dense channel connections between input and output into $G$ groups. By executing convolution in each group separately, both memory and computational overhead can be reduced by $G$ times. Meanwhile, different from the above methods that use handcrafted models, neural architecture search can automatically find lightweight CNNs with high accuracy from a pool of network candidates~\cite{lee2021hardware,elsken2019nassurvey} .

\vspace{0.3em}
\noindent \textbf{Network sparsification.} The works under this paradigm involve network pruning and low-rank decomposition to make the network more sparse. Pruning approaches aim to identify and prune the redundant parts of existing networks instead of designing a new network from scratch. With different degrees of pruning granularity, network pruning can be categorized as unstructured pruning~\cite{han2015deepcompression} and structured pruning~\cite{he2017channelpruning}. The former approaches like DeepCompression~\cite{han2015deepcompression} prune individual network weights, leading to irregular model structures and limited practical accelerations~\cite{wen2016structuredpruning}. Therefore, structured pruning was developed to prune the whole filters or channels~\cite{he2017channelpruning,luo2017thinet,li2020hinge,li2020dhp,liu2019metapruning}. Structured pruning is more hardware friendly and can be implemented with most deep learning software frameworks. 
Similarly, based on existing networks, tensor decomposition factorizes fully connected layers or convolutional layers with low-rank expressions~\cite{ouerfelli2022random,pmlr-v162-wang22ar}.

\vspace{0.3em}
\noindent \textbf{Network quantization.}
Quantization methods~\cite{zhou2016dorefa,lin2017abcnet,liu2018birealnet,jung2019quantization1,gholami2021quantizationsurvey} can aggressively keep network activations and parameters in low-bit, as the multiplicative operations between low-bit values are much more efficient than the full-precision counterparts. Specifically, a quantizer is needed in the methods to receive real-valued numbers and assign them to a countable set of value points. Our work is mostly related to network binarization~\cite{courbariaux2016bnn} in this paradigm, which is the extreme case that only utilizes binary values. A more detailed review will be given in a separate subsection.

\vspace{0.3em}
\noindent \textbf{Knowledge distillation.}
To increase the accuracy of a lightweight model, an effective way is to use the predictions or intermediate outputs of a stronger model, usually with more complex structure and higher accuracy, as the supervision to guide the training of the target model~\cite{touvron2021deit,gou2021knowledge,tung2019similaritydistill}. The technology is called knowledge distillation~\cite{hinton2015distilling} and can be seamlessly combined with other efficient representation learning methods.

\vspace{0.3em}
\noindent \textbf{Integrating traditional operators to convolution.}
Motivated by LBP~\cite{ojala2002lbp}, central difference convolution (CDC)~\cite{yu2020cdc,yu2020fas,yu2021dual,yu2021searching} and local binary convolution (LBC)~\cite{juefei2017lbc} were proposed for efficient representation learning.
Specifically, LBC uses a set of predefined sparse binary filters to extend LBP to CNN to reduce the network complexity. CDC further uses learnable weights to capture image gradient information for robust face anti-spoofing. Similarly, the proposed PDC also uses learnable filters but is more general and flexible than CDC to capture richer high-order statistics for general vision representation. In addition, inspired by Gabor descriptors, Gabor convolution encodes the orientation and scale information in the convolution kernels~\cite{luan2018gabor}. Different from Gabor convolution which multiplies kernels with a group of Gabor filters, PDC is more compact without any auxiliary traditional feature filters.

\subsection{Related Applications}

\noindent \textcolor{black}{\textbf{Edge detection.}\quad
Edge detection has been a longstanding and fundamental low-level problem in computer vision~\cite{canny1986computational}. Edges and object boundaries play an important role in various high-level computer vision tasks such as object recognition and detection \cite{liu2020deep,ferrari2007groups}, object proposal generation \cite{cheng2014bing,uijlings2013selective}, image editing \cite{elder1998imageediting}, and image segmentation \cite{muthukrishnan2011edgeimageseg,bertasius2016semantic}. 
}

\textcolor{black}{Early deep learning based edge detection models
construct CNN architectures as classifiers to predict the edge probability of an input image patch~\cite{bertasius2015deepedge,shen2015deepcontour,bertasius2015hfl}.  
Building on top of fully convolutional networks~\cite{long2015fully}, HED~\cite{xie2017holistically} performs end-to-end edge detection by leveraging multilevel image features with rich hierarchical information guided by deep supervision, and achieves state-of-the-art performance. Other similar works include~\cite{yang2016cedn,kokkinos2015deepboundary,maninis2016cob,wang2017ced,xu2018amhnet,liu2019richer,deng2018lpcb,he2019bidirectional}.
}

\textcolor{black}{
Recently, efforts have been made to design lightweight architectures for efficient edge detection~\cite{wibisono2020fined,wibisono2020traditional,poma2020dense}. Some of them may not need a pretrained network based on large-scale dataset~\cite{poma2020dense}. Although compact and fast, the detection accuracies with these networks are unsatisfactory. Different to the existing approaches, our PiDiNet owns the following three properties at the same time: 1) its prediction accuracy is on par with or superior to the existing state-of-the-art methods, 2) running at about 100 FPS on GPU with no more than 1M parameters, 3) can be trained from scratch without ImageNet pretraining.
}

\vspace{0.3em}
\noindent \textcolor{black}{\textbf{BCNNs on Object Recognition.} \quad
The research on BCNNs dates back to the pioneering work of Hubara \emph{et al.}~\cite{courbariaux2016bnn}, where the authors built a CNN with the activations and weights constrained to \{-1, +1\}. The binarization significantly helps to reduce the memory consumption of a CNN since its weights are merely stored in 1-bit. Then, Rastegari \emph{et al.}~\cite{rastegari2016xnor} further demonstrated that the inference speed of a BCNN can be 59 times faster than its full-precision counterpart on the CPU, as the matrix multiplication between 1-bit activations and 1-bit weights can be executed efficiently via the bit-wise \emph{XNOR-Count} operations. However, BCNNs suffer from 1) gradient mismatch problem during training due to the use of the straight through estimator (STE) for making the BCNNs trainable~\cite{courbariaux2016bnn}, 2) large quantization error after activation binarization, and 3) limited network capacity~\cite{liu2018birealnet}. 
In the last decades, the community has witnessed great progress in the development of BCNNs.}

\begin{figure*}[t!]
    \centering
    \includegraphics[width=1\linewidth]{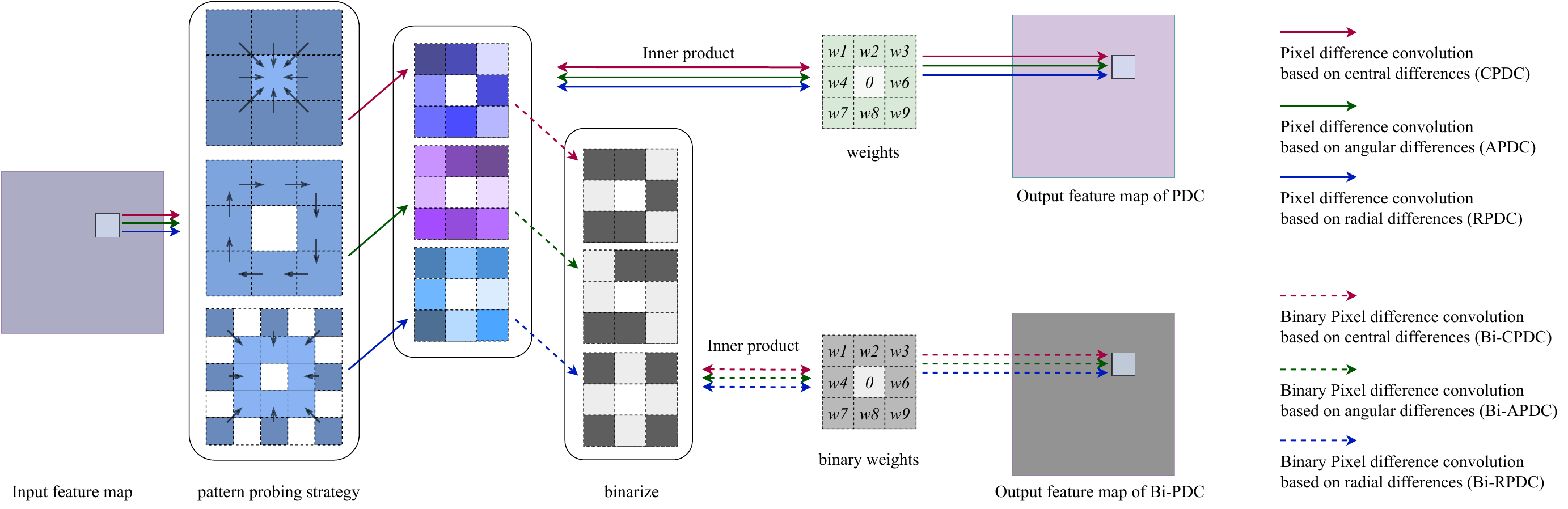}
    \caption{Three instances of pixel difference convolution derived from LBP and ELBP descriptors~\cite{liu2011sorted, liu2012extended, su2019bird}. More instances can be obtained by incorporating other probing strategy of the pixel pairs.}
    \label{fig:pdc}
\end{figure*}

\textcolor{black}{
The first problem has been alleviated from different perspectives. For example, elaborate functions were designed to give a more precise approximation of the Sign function~\cite{liu2018birealnet,qin2020forward,xu2021frequencydomain}, hyper auxiliary modules were developed to generate the binary weights such that the gradients of binary weights can be distributed in multiple paths~\cite{he2020proxybnn,wang2021gradientmatters}, and specific regularizations on activations were introduced to better guide the gradient calculation~\cite{wang2020sparsityinducing,ding2019regularizingactivations}. The second issue can be alleviated by introducing novel scaling factors, which can be pre-computed~\cite{rastegari2016xnor} or learned~\cite{bulat2019xnorpp,DBLP:conf/bmvc/bnn++,martinez2020realtobinary}. In additional, novel optimization or training methods are also helpful to minimize the quantization error~\cite{zhao2022bonn,Mingbao2020rbnn,Xu_2021_recu}. Finally, to enhance the model capacity of BCNNs, researchers have developed many effective methods like preserving the pre-binarization activations using skip connections~\cite{liu2018birealnet,liu2020reactnet}, expanding the architectures in width or branches~\cite{bulat2021highcapacityexpert,liu2019circulant,lin2017abcnet,zhu2019binaryensemble}, and modeling contextual dependencies with the help of MLPs~\cite{xing2022bcdnet}. Besides, automatically searching novel network structures is another feasible way~\cite{bulat2020bats,bulat2021highcapacityexpert}.
}

\section{PDC: Pixel Difference Convolution}
\label{sec:pdc}

In this section, we start with a preliminary introduction of LBP and then present our pixel difference convolution in details. The overall skeleton of PDC can be seen in \cref{fig:pdc}.

\subsection{Preliminaries on LBP}

\begin{figure}[t!]
    \centering
    \includegraphics[width=\linewidth]{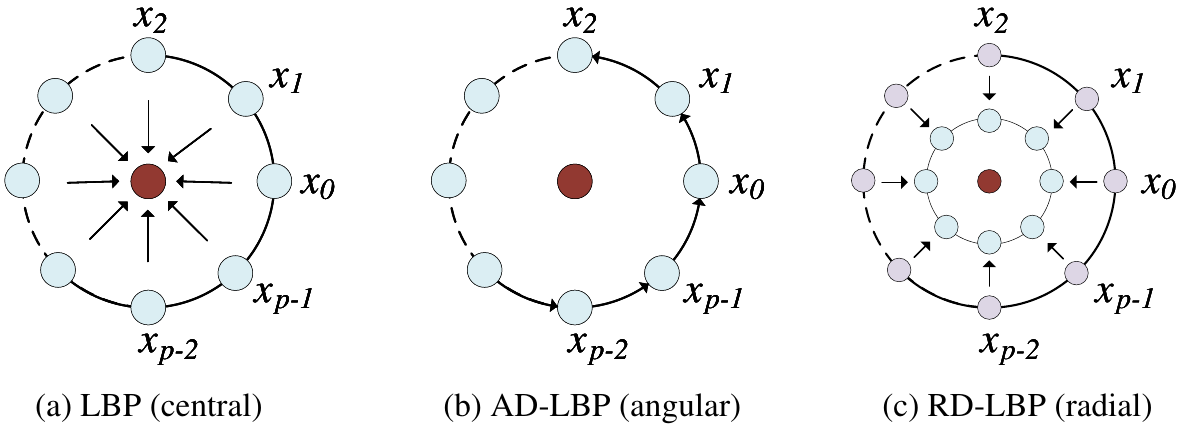}
    \caption{LBP and its variants with different probing strategies.}
    \label{fig:lbp}
\end{figure}

LBP descriptors were firstly introduced by Ojala \emph{et.al.}~\cite{ojala2002multiresolution} to encode pixel-wise information in textured images. 
Specifically, an input image is probed locally by sampling the values from the neighborhood. As shown in \cref{fig:lbp} (a), for a certain pixel $x_c$, the values from neighboring locations $\{x_0, x_1, ..., x_{p-1}\}$ spaced equidistantly around a circle are extracted to generate a binary code composed of 0 and 1, by comparing each of those values with the central value $x_c$. That is, the neighboring values greater than or equal to $x_c$ are associated with 1, otherwise with 0. 0 and 1 are read anticlockwise from the starting point $x_0$ to the ending point $x_{p-1}$, leading to a $p$-length binary code as a descriptive local pattern. 

To simplify the binary code, Ojala \etal~\cite{ojala2002multiresolution} proposed ``uniform'' patterns where the number of spatial transitions (bitwise 0/1 changes) in the code is at most 4. The ``uniform'' patterns show stronger noise-robustness and are widely used to capture general micro-structures such as bright and dark spots, flat areas, and edges. In ELBP~\cite{liu2012extended,su2019bird}, more sophisticated encoding strategies were proposed to enhance the representation ability of LBP descriptors.  For example, as shown in \cref{fig:lbp} (b-c), AD-LBP and RD-LBP compare neighboring pixels on the circular and radii grid to probe the pattern of intensity changes along the angular and radial directions, respectively.

\subsection{PDC: Integrating LBP into Convolution}

Different from vanilla convolution, our PDC incorporates the calculation of pixel differences when conducting convolution operation. 
The formulations of vanilla convolution and PDC can be written as:
\begin{align}
    y &= f(\pmb{x}, \pmb{\theta}) = \sum_{i=1}^{k\times k}w_{i}\cdot x_{i}, \;\;\;\;\;\;\; \text{(vanilla convolution)} \label{eq:vc} \\
    y &= f(\Delta\pmb{x}, \pmb{\theta}) = \sum_{(x_i, x_i')\in \pmb{\mathcal{P}}}w_{i}\cdot (x_i - x_i'), \;\;\;\;\;\;\, \text{(PDC)} \label{eq: pdc}
\end{align}
where $\pmb{x}$ and $\pmb{\theta}$ represent input in a local region and filter weights, respectively.  $x_i$ and $x_i'$ are the input pixels, and $w_i$ is the weight in the $k \times k$ convolutional kernel. $\pmb{\mathcal{P}} = \{(x_1, x_1'), (x_2, x_2'), ..., (x_m, x_m')\}$ is the set of pixel pairs picked from the current local region, and $m\le k\times k$. 

To better capture diverse micro-structural patterns, pixel pairs can be selected according to probing strategies inspired by different traditional feature descriptors. Here, LBP and ELBP~\cite{ojala2002lbp,liu2012extended,su2019bird} are adopted to encode pixel relations from varying directions (angular and radial). By integrating LBP and ELBP into convolution, we derive three types of PDC instances as shown in \cref{fig:pdc}, denoted as central PDC (CPDC), angular PDC (APDC), and radial PDC (RPDC), respectively. For example, for APDC with a $3\times 3$ kernel, 8 pairs are first selected in the angular direction within the $3\times 3$ local region (thus $m=8$). Then,  pixel differences between these pixel pairs are convolved with 
convolutional kernel to calculate the output feature map.

\begin{figure}[t!]
    \centering
    \includegraphics[width=0.8\linewidth]{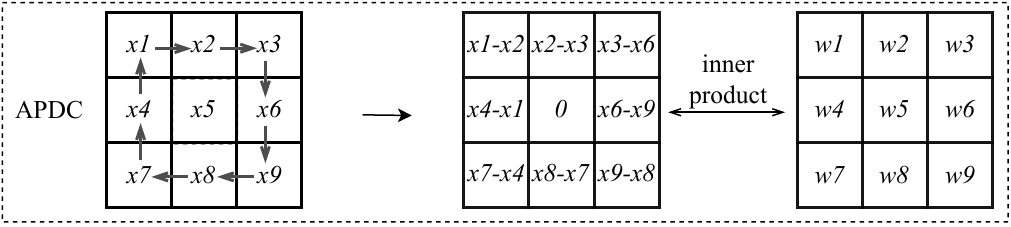}
    \caption{Selection of pixel pairs and convolution in APDC.}
    \label{fig:apdc}
\end{figure}

\vspace{0.3em}
\noindent \textbf{Re-parameterization Strategy.}
One may notice that the computational cost and memory footprint of PDC (\cref{eq: pdc}) are doubled as compared with the vanilla convolution (\cref{eq:vc}). 
To remedy this, we propose a re-parameterization strategy by calculating kernel differences instead of pixel differences for efficient implementation. 
For example, as illustrated in \cref{fig:apdc}, the output of an APDC layer can be re-written as:
\begin{align}
    y &= w_{1}\cdot (x_1 - x_2) + w_2\cdot (x_2 - x_3)+w_3\cdot (x_3 - x_6) + ... \nonumber\\
    &=(w_1 - w_4)\cdot x_1 + (w_2 - w_1)\cdot x_2 + (w_3 - w_2)\cdot x_3 + ...\nonumber\\
    &=\hat{w}_1\cdot x_1 + \hat{w}_2\cdot x_2 + \hat{w}_3\cdot x_3 + ... \nonumber\\
    &=\sum \hat{w}_i\cdot x_i,
\end{align}
where $\{\hat{w}_i\}$ are the re-parameterized kernel weights. During the training phase, since the additional overhead for calculating kernel differences is negligible, our PDC is as efficient as vanilla convolution in terms of both computational cost and memory footprint. After training, we directly save the re-parameterized weights $\{\hat{w}_i\}$ as our model weights, as the original kernel weights $\{w_i\}$ are no longer needed..

\section{Binary Pixel Difference Convolution}
\label{sec:Bi-PDC}

\textcolor{black}{
Current PDC is still in full-precision, which may prevent it from deploying on devices where the memory and computing resources are strictly constrained. As mentioned in the introduction, PDC is compatible with network binarization to obtain further efficiency gains. To complete the proposed PDC, in this section, we introduce its binary version, Bi-PDC, in which the calculation of convolution can be implemented by the efficient bit-wise operations and the memory consumption is significantly reduced. 
}

\subsection{Preliminaries on Binary Convolution}
Generally, convolution operations are conducted with full-precision 32-bit numbers. To further improve the computational and memory efficiency of the convolution operation, binary convolution \cite{courbariaux2016bnn} has been developed, with both activations and weights being binarized to 1-bit values (\ie \{-1, +1\}). Within the binary convolution, the weights and activations are first binarized using a Sign function:
\begin{equation}
	\text{Sign}(x) = \begin{cases}
		+1\;\;\text{if}\;\; x >= 0,\\
		-1\;\;\text{otherwise}.
	\end{cases}
	\label{eq:sign}
\end{equation}
Then, the inner products between binarized activations and binary weights are computed to obtain the output features. By representing activations and weights using 1-bit values, binary convolution can be efficiently executed using bit-wise \emph{XNOR-Count} operations~\cite{rastegari2016xnor}.

\begin{figure}[t!]
    \centering
    \includegraphics[width=\linewidth]{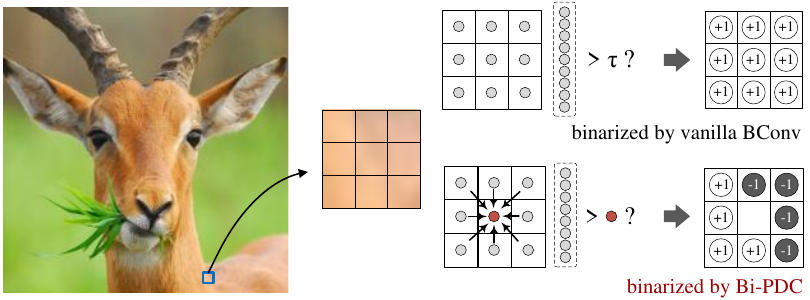}
    \caption{During vanilla BConv, pixels in certain local regions (for example, a pixel patch from the antelope skin) are all binarized to the same value, since the threshold was optimized for the whole input rather than a particular patch area, hence eliminating the valuable micro-structural information. By using a neighboring pixel (the central pixel in this example of Bi-CPDC) as the threshold, in contrast, Bi-PDC effectively preserves such information without any extra parameters introduced.}
    \label{fig:micro}
\end{figure}

\begin{figure*}[t!]
    \centering
    \includegraphics[width=\linewidth]{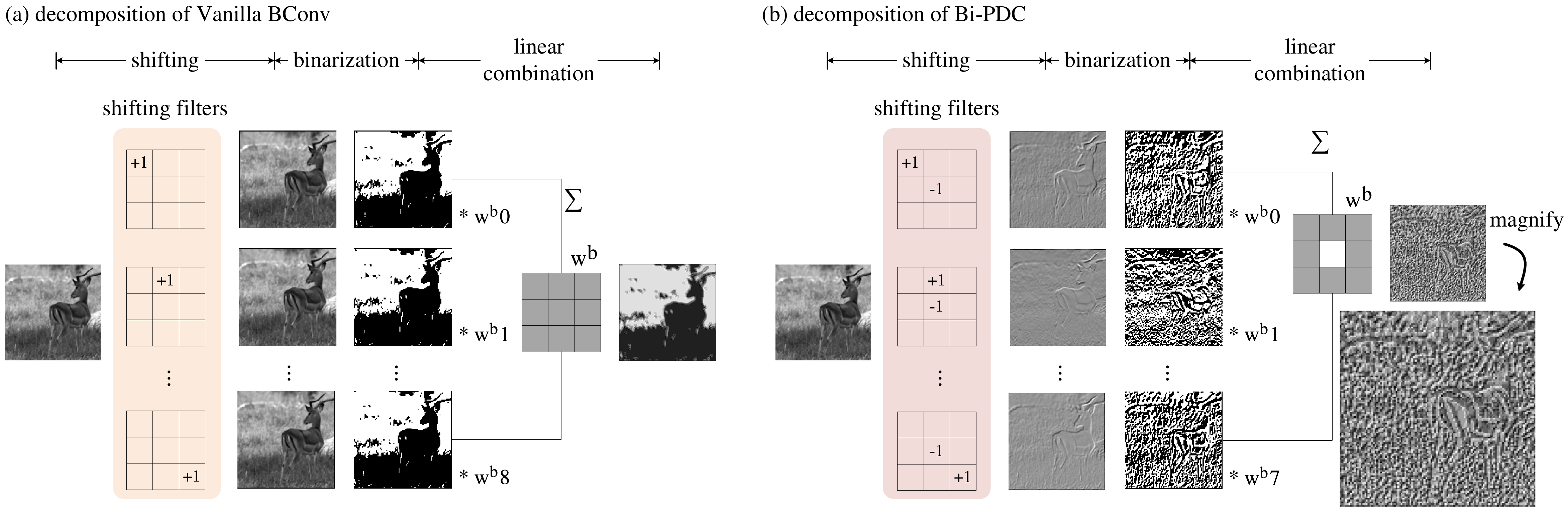}
    \caption{Binary convolutional process can be equivalently decomposed into the following procedures: the input is shifted according to the certain shifting filters $\rightarrow$ binarization on these shifted version $\rightarrow$ linear combination of the binarized maps with the coefficients from the binary weights to generate the output map. Vanilla BConv and Bi-PDC differ in the related shifting filters, leading to different behaviors in information extraction.}
    \label{fig:decomposition}
\end{figure*}

\begin{figure}[t!]
    \centering
    \includegraphics[width=\linewidth]{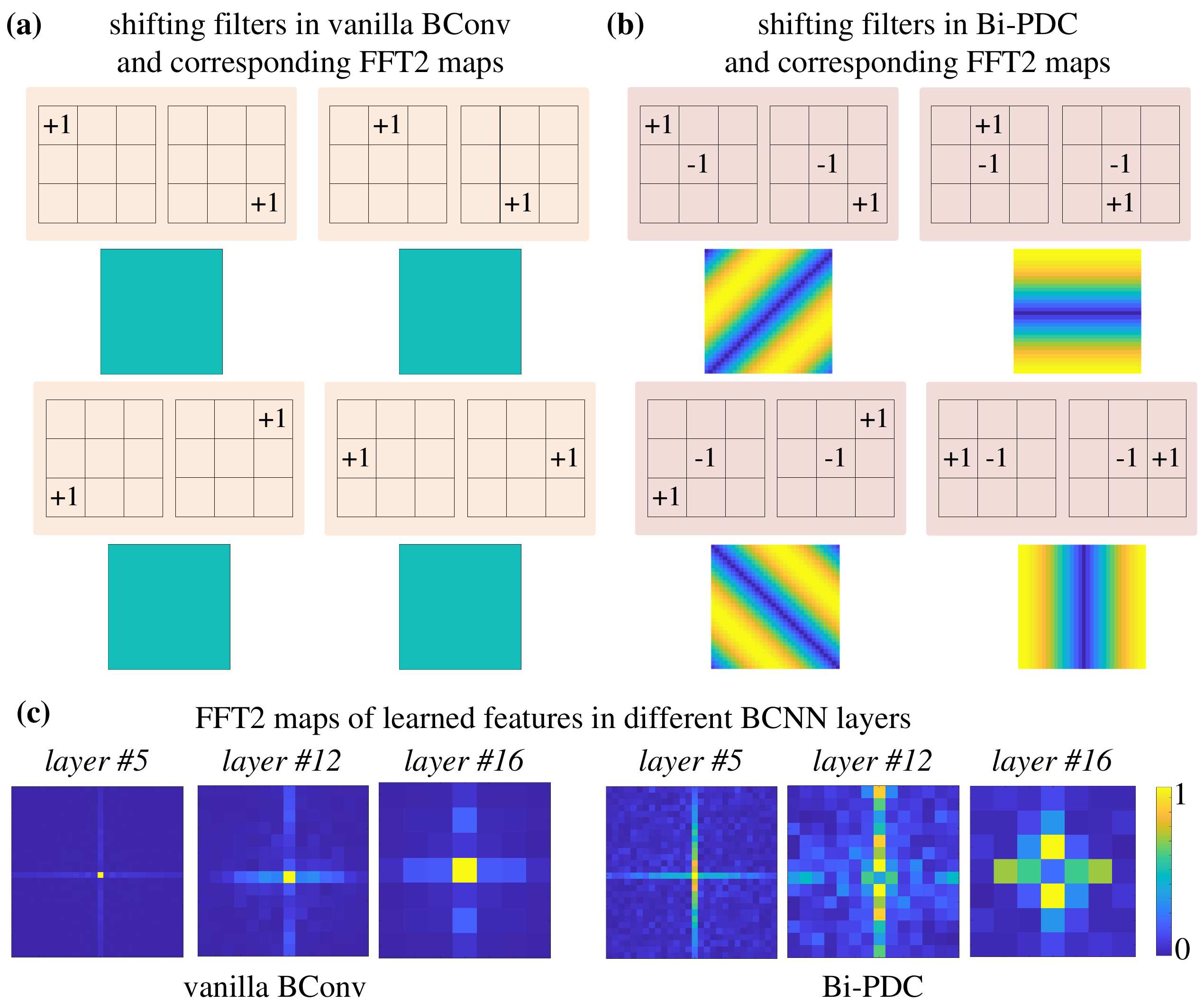}
    \caption{FFT2 maps of shifting filters and actual feature maps.}
    \label{fig:frequency}
\end{figure}

\subsection{Formulations of Bi-PDC}

Similar with our observation in full-precision CNNs, vanilla binary convolution in BCNNs cannot well capture high-order information, as illustrated in \cref{fig:figure2}(b). Consequently, we further extend the idea of our PDC to binary convolutions (BConvs) to develop binary pixel difference convolution (Bi-PDC). 
The formulations of vanilla BConv and our Bi-PDC can be written as:
\begin{align}
    y &= f(\pmb{x}^b, \pmb{\theta}^b) = \sum_{i=1}^{k\times k}w_{i}^b\cdot \text{Sign}(x_{i} - \tau), \;\;\;\;\;\;\; \text{(vanilla BConv)} \label{eq:vbc} \\
    y &= f((\Delta\pmb{x})^b, \pmb{\theta}^b) = \sum_{(x_i, x_i')\in \pmb{\mathcal{P}}}w_{i}^b\cdot \text{Sign}(x_i - x_i'), \;\;\;\;\;\;\, \text{(Bi-PDC)} \label{eq:bipdc}
\end{align}
where $\pmb{x}^b$ and $\pmb{\theta}^b$ represents the binarized input and binary weights, respectively. $\tau$ is the threshold in vanilla BConv to binarize the activations. 
Similarly, Bi-CPDC, Bi-APDC, and Bi-RPDC can be derived using different probing strategies.

Generally, previous BCNNs~\cite{rastegari2016xnor,liu2020reactnet,zhang2022dynamicthreshold} adopted a shared threshold $\tau$ over the whole image during binarization without considering local content variations, leading to the irreversible loss of high-order image details. As illustrated in \cref{fig:micro}, when a global threshold is used, pixels within the $3\times 3$ region are all binarized to +1 with local texture details being discarded. In contrast, our Bi-PDC uses dynamic thresholds conditioned on neighboring pixels for binarization, which preserves the high-order texture details.

\section{Frequency Domain Interpretation}
\label{sec:theoretical}

In this section, we give an interpretation to our PDC from a perspective of frequency domain. Without loss of generality, only Bi-PDC is used for analyses and the conclusions also hold for PDC. 

For the convenience of analyses, we decompose binary convolutions into three steps, as shown in \cref{fig:decomposition}. First, the input image (feature map) is convolved with corresponding shifting filters\footnote{It should be noted that these shifting filters are only used for analysis and do not necessarily exist in implementation.}. Second, shifted images (feature maps) are binarized. Third, the binarized results are aggregated using the binary kernel weights as coefficients, generating the output feature map.
Since shifting filters in vanilla BConv essentially move the input image with particular offsets, zeroth-order information (\ie absolute intensities) is preserved. In contrast, the shifting filters in Bi-PDC calculate local intensity variations along different directions like LBP does. As a result, Bi-PDC pays more attention to high-frequency components.

According to the convolution theorem in signal processing, convolution in the spatial domain is equivalent to pointwise multiplication in the frequency domain:
\begin{equation}
     \mathbf{Y} = \mathbf{X} \star \mathcal{S} = {\rm IFFT2}(\text{FFT2}(\mathbf{X}) \cdot {\rm FFT2}(\mathcal{S})),
     \label{eq:fft}
\end{equation}
where $\rm FFT2(\cdot)$ and $\rm IFFT2(\cdot)$ mean 2D fast Fourier transform and 2D inverse fast Fourier transform, $\mathcal{S}$ represents a shifting filter. 
We illustrate the FFT2 maps of the shifting filters for vanilla BConv and Bi-PDC in \cref{fig:frequency} (a-b).  
It can be seen that Bi-PDC actually employs a series of high-pass filters to extract high-frequency information. In contrast, 
shifting filters in vanilla BConv do not highlight any
frequency components.
When only vanilla BConv is adopted in BCNNs,  features are dominated by the inherent low-frequency components, drowning out the high-frequency information that is beneficial to enhancing BCNNs' representational capacity.

We further visualize the FFT2 results of feature maps in BConv and Bi-PDC averaged over 100 randomly sampled images from the ImageNet validation set (\cref{fig:frequency} (c)). It can be observed that our Bi-PDC highlights more high-frequency components as compared with the vanilla BConv.

\section{Application-specific Network Architectures}
\label{sec:application}

In this section, we apply our PDC and Bi-PDC to the semantically low-level task of edge detection and the high-level task of object recognition, respectively. Specifically, two lightweight networks are developed, namely PiDiNet and Bi-PiDiNet.

\subsection{PiDiNet for Edge Detection}
\label{sec:pidinet}

Leading CNN based edge detectors suffer from big memory storage and high computational cost with the large ImageNet pretraining backbones~\cite{xie2017holistically,liu2019richer,he2019bidirectional}, due to the fact that the annotated data available for training edge detection models is limited, and thus a well pretrained backbone is needed.

\begin{figure}[t!]
    \centering
    \includegraphics[width=\linewidth]{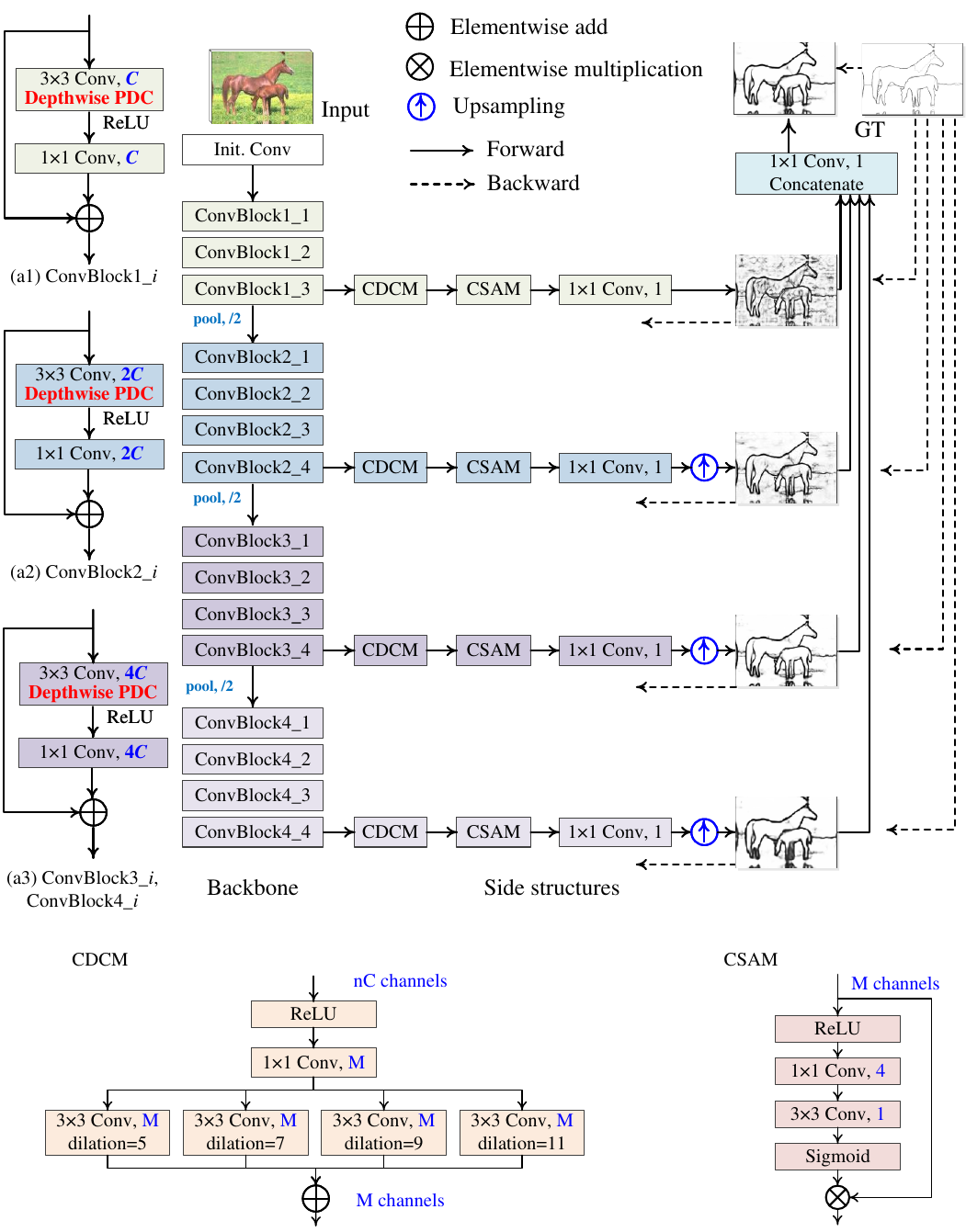}
    \caption{Network Architecture of our PiDiNet.}
    \label{fig:arch}
\end{figure}

As tried by some prior works~\cite{wibisono2020fined,poma2020dense,wibisono2020traditional}, we believe it is both necessary and feasible to solve the above inefficiencies at one time by building an architecture with small model size and high running efficiency, and can be trained from scratch using limited datasets for effective edge detection. We construct our PiDiNet with the following parts (\cref{fig:arch}).

\vspace{0.3em}
\noindent \textbf{Efficient Backbone.} \quad The building principle for the backbone is to make the structure slim with high running efficiency. To this end, we design a simple yet effective network without multi-branch lightweight structures, as shown in \cref{fig:arch}.
Specifically, the backbone has 4 stages with max pooling layers being adopted between adjacent stages for downsampling.
Each stage has 4 residual blocks (except the first stage that has an initial convolutional layer and 3 residual blocks). The residual path in each block includes a depthwise convolutional layer, a ReLU layer, and a pointwise convolutional layer sequentially.
The number of channels in each stage is reasonably small to avoid big model size ($C$, $2\times C$, $4\times C$ and $4\times C$ channels for stage 1, 2, 3, and 4 respectively).

\vspace{0.3em}
\noindent \textbf{Efficient Side Structures.} \quad
To learn rich hierarchical edge representation, we also use the side structure as in~\cite{xie2017holistically} to generate an edge map from each stage respectively. Meanwhile, a side loss is adopted to provide deep supervision~\cite{xie2017holistically}.
To refine the feature maps, beginning from the end of each stage, we firstly build a compact dilation convolution based module (CDCM) to enrich multi-scale edge information. Our CDCM takes the input with $n\times C$ channels, and produces $M$ ($M < C$) channels in the output to relieve the computation overhead, followed by a compact spatial attention module (CSAM) to eliminate the background noise. After that, a $1\times 1$ convolutional layer further reduces the feature volume to a single channel map, which is then upsampled to the original size. Finally, we use the Sigmoid function to create the edge map. The final edge map, which is used for testing, is created by fusing the 4 single channel feature maps with a concatenation, a $1\times 1$ convolutional layer and a Sigmoid function. 

\textcolor{black}{The $3\times$3 depthwise convolutional layers in the backbone blocks are constructed using our PDC. Ablation study of how to configure different PDC instances are detailed in \cref{sec:ablation-edge}}. Besides, batch normalization layers are not adopted since the resolutions of the training images are not uniform.

\vspace{0.3em}
\noindent \textbf{Loss Function.} \quad We adopt the annotator-robust loss function proposed in~\cite{liu2019richer} for each generated edge map (including the final edge map). For the $i$th pixel in the $j$th edge map with value $p_i^j$, the loss is calculated as:
\begin{equation}
    l_i^j = \begin{cases} 
    -\alpha\cdot \log (1 - p_i^j) &\text{if } y_i  = 0 \\
    0 &\text{if } 0 < y_i < \eta \\
    -\beta\cdot \log p_i^j &\text{otherwise},
    \end{cases}
\end{equation}
where $y_i$ is the ground truth edge probability, $\eta$ is a pre-defined threshold, $\beta$ is the percentage of negative pixel samples and $\alpha = \lambda\cdot (1-\beta)$.
Pixels marked with fewer than $\eta$ annotators are not included in the loss to avoid confusion.
Overall, the total loss is $L = \sum_{i,j}l_i^j$.

\subsection{Bi-PiDiNet for Object Recognition}
\label{sec:bipidinet}
 
\textcolor{black}{Following most of the BCNN approaches, we choose the general object recognition task to validate the effectiveness of our Bi-PDC.
It is commonly known that binarized networks suffer from limited accuracy.
Therefore, we introduce our Bi-PiDiNet that can flexibly integrate our Bi-PDC with vanilla BConv to exploit complementary high-order information for higher accuracy.
}

\begin{figure}[t!]
    \centering
    \includegraphics[width=\linewidth]{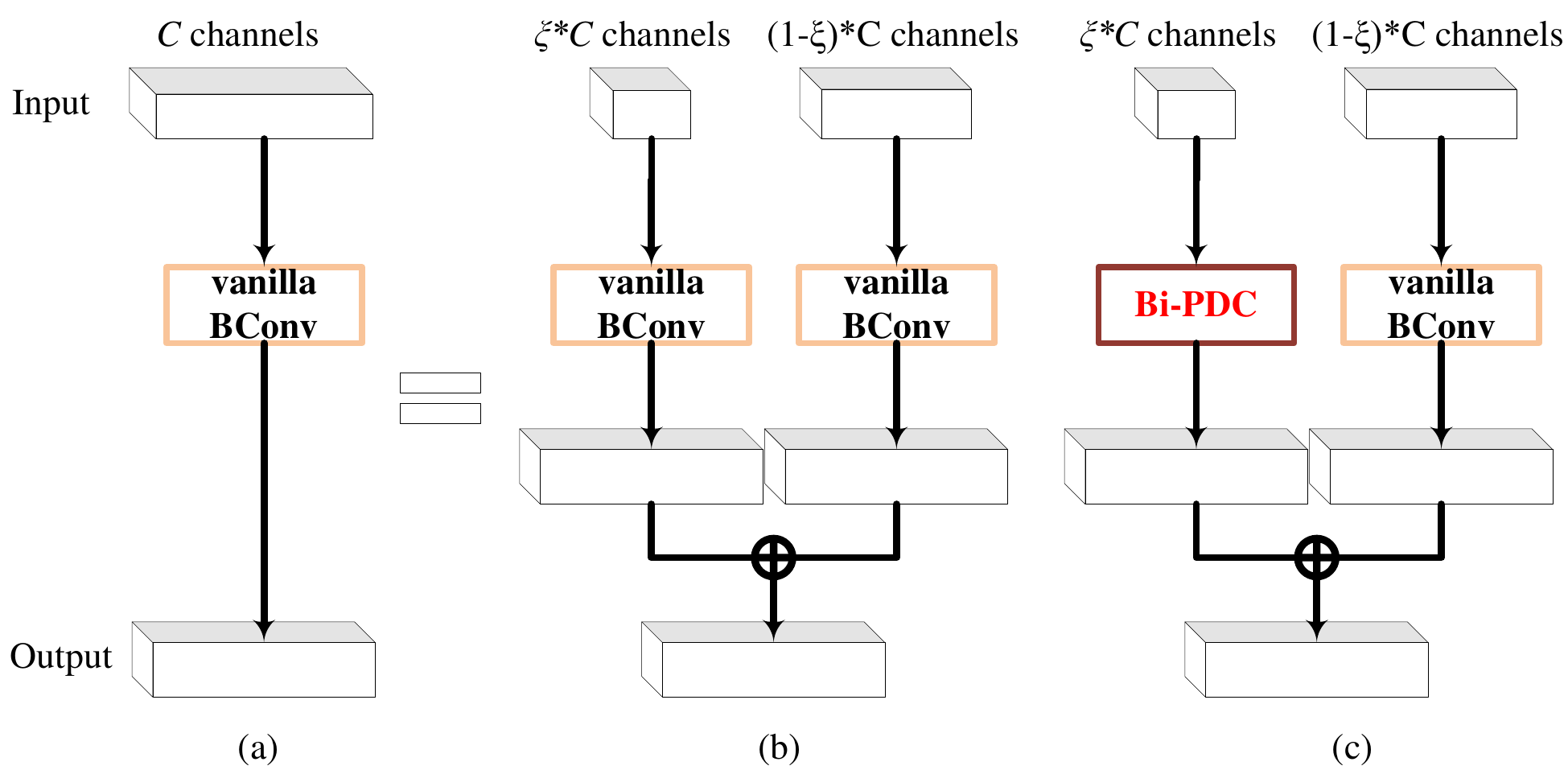}
    \caption{(a): A vanilla binary convolutional layer; (b): an equivalent decomposition of (a), where the channels are split from the input end; (c): a binary convolutional layer with $\xi$ proportion of channels fed to Bi-PDC. (a) and (c) share the same network complexity.}
    \label{fig:fusion}
\end{figure}

\vspace{0.3em}
\noindent \textbf{Layer-level: Both zero-order and high-order information matter.}
\quad A vanilla binary convolutional layer can be written as:

\begin{equation}
    \label{eq:vanilla}
    \mathbf{Y} = f(\text{BConv}(\mathbf{X})),
\end{equation}
where $f$ represents batch normalization (BN)~\cite{IoffeS15batchnorm} and ReLU layers, $\mathbf{X}$ and $\mathbf{Y}$ are input and output feature maps. 
As analyzed in Section~\ref{sec:ablation-image}, our Bi-PDC can provide complementary information to vanilla BConv. Therefore, we are motivated to combine these two types of convolutions to enhance the representation capacity of the convolutional layer. Specifically, as illustrated in \cref{fig:fusion} (c), input features are first split into two parts and fed to vanilla BConv and our Bi-PDC, respectively. Then, the resultant features are aggregated though a summation to produce the output features. The formulation of our convolutional layer can be obtained as:
\begin{align}
    \mathbf{Y} &= f(\text{Bi-PDC}(\mathbf{X}[:, :(\xi\cdot C), :, :]))\nonumber \\
    &+ f(\text{BConv}(\mathbf{X}[:, (\xi\cdot C):, :, :])),
    \label{eq:hybrid}
\end{align}
where $C$ is the number of channels in $\mathbf{X}$ and $\xi$ determines the proportion of channels fed to Bi-PDC. This structure degrades to a vanilla binary convolutional layer when $\xi=0$. As $\xi$ increases, more channels are processed by Bi-PDC such that more high-order information can be captured. The effect of  $\xi$ is analyzed in Section~\ref{sec:ablation-image}.

\begin{figure}[t!]
    \centering
    %\vspace{-0.1in}
    \includegraphics[width=0.8\linewidth]{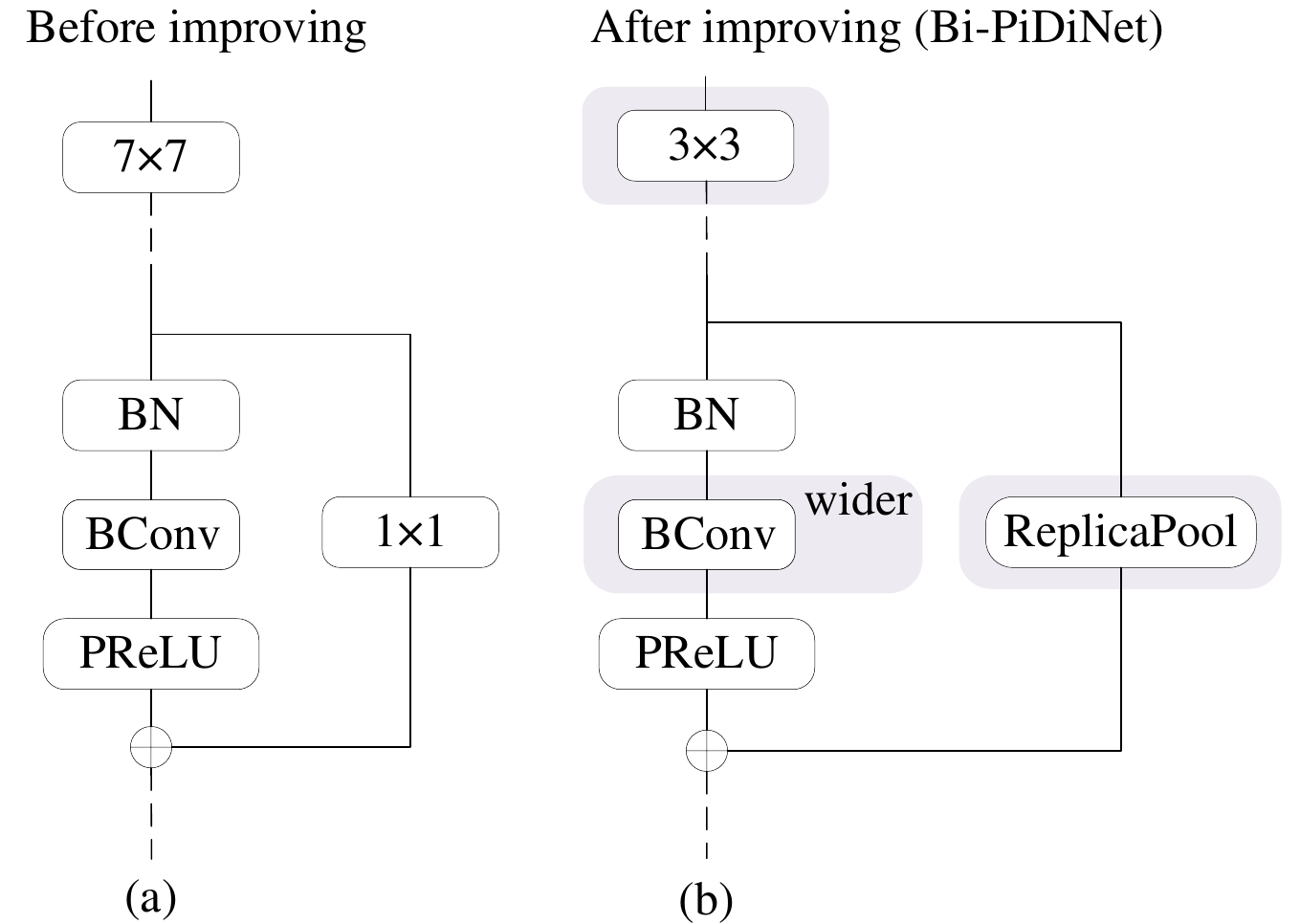}
    \caption{Illustration of Bi-PiDiNet. Here we only show the initial layer and reduction blocks (for normal blocks, no $1\times 1$ convolutional layer is needed in the shortcut).}
    \label{fig:bipidinet}
\end{figure}

\begin{algorithm}
\caption{Parameter-free ReplicaPool Layer}\label{alg:replicapool}
\begin{algorithmic}[1]
\Require Input $\mathbf{X}\in\mathbb{R}^{B\times C\times H\times W}$, output $\mathbf{Y}\in\mathbb{R}^{B\times C'\times H'\times W'}$, where the dimensions of $\mathbf{X}$ or $\mathbf{Y}$ represent batch size, number of channels, height, and width, respectively. $C' = M\times C + C/N$, $M$ and $N$ are positive integers, $H' = H/2$, $W' = W/2$
\State $\mathbf{X} \gets \text{AvgPool}(\mathbf{X})$
\Comment{Downsample $\mathbf{X}$ by half in height and width dimensions}
\State $\mathbf{Y}_i \gets \mathbf{X}$ for $i\in {1,2,...,M}$
\State $\{\mathbf{X}_i|i=1,2,...,N\} \gets \text{split}(\mathbf{X})$
\Comment{Split $\mathbf{X}$ into $N$ even segments in channel dimension}
\State $\mathbf{Y}_{M+1} = \frac{1}{N}\sum_{i=1}^{N}\mathbf{X}_i$
\State $\mathbf{Y} = \text{Concate}(\{\mathbf{Y}_i|i=1,2,...,M+1\})$
\Comment{Concatenate $\{\mathbf{Y}_i\}$ in channel dimension}
\end{algorithmic}
\end{algorithm}

\vspace{0.3em}
\noindent \textbf{Network-level: A more powerful binary backbone matters.}
\quad To construct powerful binary backbones, we start by incorporating the recent insights on building high accuracy BCNNs based on ResNet architectures~\cite{he2016residual}: Firstly, we add an additional shortcut~\cite{liu2018birealnet} in each convolutional block to facilitate high capacity information flow. Secondly,  we adopt PReLU~\cite{he2015delving} activation function as in~\cite{martinez2020realtobinary} for better training. Finally, we use identity mappings to ease training and improve generalization as studied in~\cite{he2016identity}.

Next, we further improve our structure by considering the actual computational cost of the binary model (\cref{fig:bipidinet}). Specifically, following~\cite{liu2020reactnet,zhao2022bonn}, we use the number of FLOPs (full-precision operations), BOPs (binary operations), and OPs (the total operations \#OPs = \#FLOPs + $1/64\times$\#BOPs) to compare the computational costs of BCNNs. To make our network more lightweight, a general rule is to reduce the usage of full-precision layers or replace them with binary counterparts. Precisely, we observed that the $7\times 7$ full-precision initial layer and the $1\times 1$ convolutional layers in the downsampling shortcuts in basic BCNN architectures~\cite{zhao2022bonn,Xu_2021_recu,han2020noisysup,xu2021frequencydomain} consume the vast majority of total computations (about 80\% of total OPs). Therefore, we firstly slim the initial layer to a $3\times 3$ convolutional layer, and propose a parameter-free pooling layer (ReplicaPool) as illustrated in Algorithm~\ref{alg:replicapool}. ReplicaPool allows us to expand the channels in a more efficient way that avoids the use of the full-precision $1\times 1$ convolutions. Finally, to compensate for the representational capacity reduced by these modifications, we widen the binary convolutional layers but maintain a small consumption of OPs for the whole structure, as doing so only introduces extra BOPs, which are far cheaper than FLOPs.

\section{Experiments}
\label{sec:experiments}

\subsection{PiDiNet on Edge Detection}

\noindent \textbf{Datasets.} \quad We evaluated the proposed PiDiNet on three widely used datasets, namely, BSDS500~\cite{arbelaez2010bsds}, NYUD~\cite{shi2000nyud}, and Multicue~\cite{mely2016multicue}. 
The BSDS500 dataset consists of 200, 100, and 200 images in the training set, validation set, and test set, respectively. Each image has 4 to 9 annotators.
Like prior works~\cite{xie2017holistically,liu2019richer,he2019bidirectional},
the PASCAL VOC Context dataset~\cite{mottaghi2014voc}, which has 10K labeled images (and was augmented to 20K with flipping), was also included for training. The NYUD dataset has 1449 pairs of aligned RGB and depth images which were densely labeled. There are 381, 414, and 654 images for training, validation, and test, respectively.
The Multicue dataset is composed of 100 challenging natural scenes and each scene contains left- and right-view color sequences captured by a binocular stereo camera. The last frame of left-view sequences for each scene, which was labeled with edges and boundaries, was used in our experiments.

\vspace{0.3em}
\noindent \textbf{Evaluation Metrics.} \quad During evaluation, \emph{F-measure} values at both Optimal Dataset Scale (ODS) and Optimal Image Scale (OIS) were recorded for all datasets. Since efficiency is one of the main focuses in this paper, all the models were compared based on the evaluations from single scale images if not specified. In addition, we compare the inference speeds in FPS (frames per second) for the models. When testing on GPU, we use the actual dataset. While on CPU, we run each model on a randomly created $3\times224\times 224$ tensor 100 times during testing.

\vspace{0.3em}
\noindent \textbf{Training settings.} \quad
Instead of using ImageNet to pretrain the backbone, we randomly initialized our models and trained them from scratch for 20 epochs with Adam optimizer~\cite{kingma2014adam}. The initial learning rate was 0.005, which was decayed in a multi-step way (at epoch 10 and 16 with decaying rate 0.1). $\lambda$ was set to 1.1 for both BSDS500 and Multicue, and 1.3 for NYUD. The threshold $\eta$ was set to 0.3 for both BSDS500 and Multicue.
No $\eta$ was needed for NYUD
since the images were singly annotated. 
All experiments were conducted based on the Pytorch library~\cite{paszke2019pytorch}.

\subsubsection{Model Analyses}
\label{sec:ablation-edge}

To demonstrate the effectiveness of PDC, we conducted our ablation study on the BSDS500 dataset. If not specified, we utilized the training set mixed with the VOC dataset for training and record the metrics on the validation set. Images in the BSDS500 training set were augmented with flipping (2$\times$), scaling (3$\times$), and rotation (16$\times$).

\vspace{0.3em}
\noindent \textbf{Architecture Configuration.} \quad We can replace the vanilla convolution with PDC in any block (we also regarded the initial convolutional layer as a block in the context) in the backbone. Since there are 16 blocks, and a brute force search for the architecture configurations is not feasible, hence we only sampled some of them as shown in \cref{table:configuration} by gradually increasing the number of PDCs. 

In the first row of \cref{table:configuration}, we found replacing the vanilla convolution with PDC only in a single block can even have obvious improvement. Then we gradually added more PDC layers as shown in the second, third, and fourth row of the table. By observing the single columns, we saw more replacements with the same type of PDC may no longer give extra performance gain and instead degenerated the model. We conjecture that the PDC in the first block already obtains much gradient information from the raw image, and excessive use of PDC may even cause the model fail to preserve useful information. The extreme case is that when all the blocks were configured with PDC as shown in the fourth row, the performance became worse than that of the baseline. The best configuration is `[CARV]$\times$4', which means combing the 4 types of convolutions (including the vanilla convolution) sequentially in each stage, as different types of PDC capture the gradient information in different encoding directions. A more straightforward comparison is between `[CARV]$\times$4' and the configurations in the third row of the table, where PDCs of the same type were used. The variation of PDCs in `[CARV]$\times$4' led to better performance. Therefore, we used this configuration in the following experiments. 

To further demonstrate the superiority of PiDiNet over the baseline, which only uses the vanilla convolution, we gave more comparisons as shown in \cref{table:morecomparison}. It constantly proves that PDC configured architectures outperform the corresponding vanilla convolution configured architectures.

\begin{table}[t!]
\caption{Configurations of PiDiNet. `C', `A', `R' and `V' indicate CPDC, APDC, RPDC and vanilla convolution respectively. `$\times$n' means repeating the pattern for $n$ times sequentially. For example, the baseline architecture can be presented as ``[V]$\times$16'', and `C-[V]$\times$15' means using CPDC in the first block and vanilla convolutions in the later blocks.}
\begin{center}
\setlength{\tabcolsep}{0.04\linewidth}
\renewcommand{\arraystretch}{1.2}
\resizebox*{\linewidth}{!}{
\begin{tabular}{llll}
\toprule
Architecture & C-[V]$\times$15 & A-[V]$\times$15 & R-[V]$\times$15 \\
ODS / OIS & 0.775 / 0.794 & 0.774 / 0.794 & 0.774 / 0.792 \\
\midrule
Architecture & [CVVV]$\times$4 & [AVVV]$\times$4 & [RVVV]$\times$4 \\
ODS / OIS & 0.773 / 0.792 & 0.771 / 0.790 & 0.772 / 0.791 \\
\midrule
Architecture & [CCCV]$\times$4 & [AAAV]$\times$4 & [RRRV]$\times$4 \\
ODS / OIS & 0.772 / 0.791 & 0.775 / 0.793 & 0.771 / 0.787 \\
\midrule
Architecture & [C]$\times$16 & [A]$\times$16 & [R]$\times$16 \\
ODS / OIS & 0.767 / 0.786 & 0.768 / 0.786 & 0.758 / 0.777 \\
\midrule
Architecture & Baseline & \multicolumn{2}{l}{\textbf{[CARV]$\times$4 (The adopted one)}} \\
ODS / OIS & 0.772 / 0.792 & \multicolumn{2}{l}{\textbf{0.776 / 0.795}} \\
\bottomrule
\end{tabular}
}
\end{center}
\label{table:configuration}
\end{table}

\begin{table}[t!]
\caption{More comparisons between PiDiNet and the baseline architecture in multiple network scales by changing the number of channels $C$ (see \cref{fig:arch}). The models were trained only using the BSDS500 training set, and evaluated on BSDS500 validation set.}
\begin{center}
\setlength{\tabcolsep}{0.06\linewidth}
\resizebox*{\linewidth}{!}{
\begin{tabular}{lcc}
\toprule
Scale & Baseline & PidiNet-[CARV]$\times 4$ \\
\midrule
Tiny (C=20) & 0.735 / 0.752 & \textbf{0.747 / 0.764} \\
\midrule
Small (C=30) & 0.738 / 0.759 & \textbf{0.752 / 0.769} \\
\midrule
Basic (C=60) & 0.736 / 0.751 & \textbf{0.757 / 0.776} \\
\bottomrule
\end{tabular}
}
\end{center}
\label{table:morecomparison}
\end{table}

\begin{table}[t!]
\caption{Ablation on CDCM, CSAM and shortcuts.}
\begin{center}
\setlength{\tabcolsep}{0.025\linewidth}
\resizebox*{\linewidth}{!}{
\begin{tabular}{lccc|c}
\toprule
Model & CSAM & CDCM & Shortcuts & ODS / OIS \\
\midrule
PiDiNet-L (light) & \xmark & \xmark & \checkmark & 0.770 / 0.790 \\
\midrule
- & \xmark & \checkmark & \checkmark & 0.775 / 0.793 \\
\midrule
- & \checkmark & \checkmark & \xmark & 0.734 / 0.755 \\
\midrule
PiDiNet & \checkmark & \checkmark & \checkmark & \textbf{0.776 / 0.795} \\
\bottomrule
\end{tabular}
}
\end{center}
\label{table:moreablation}
\end{table}

\begin{figure}[t!]
    \centering
    \includegraphics[width=1\linewidth]{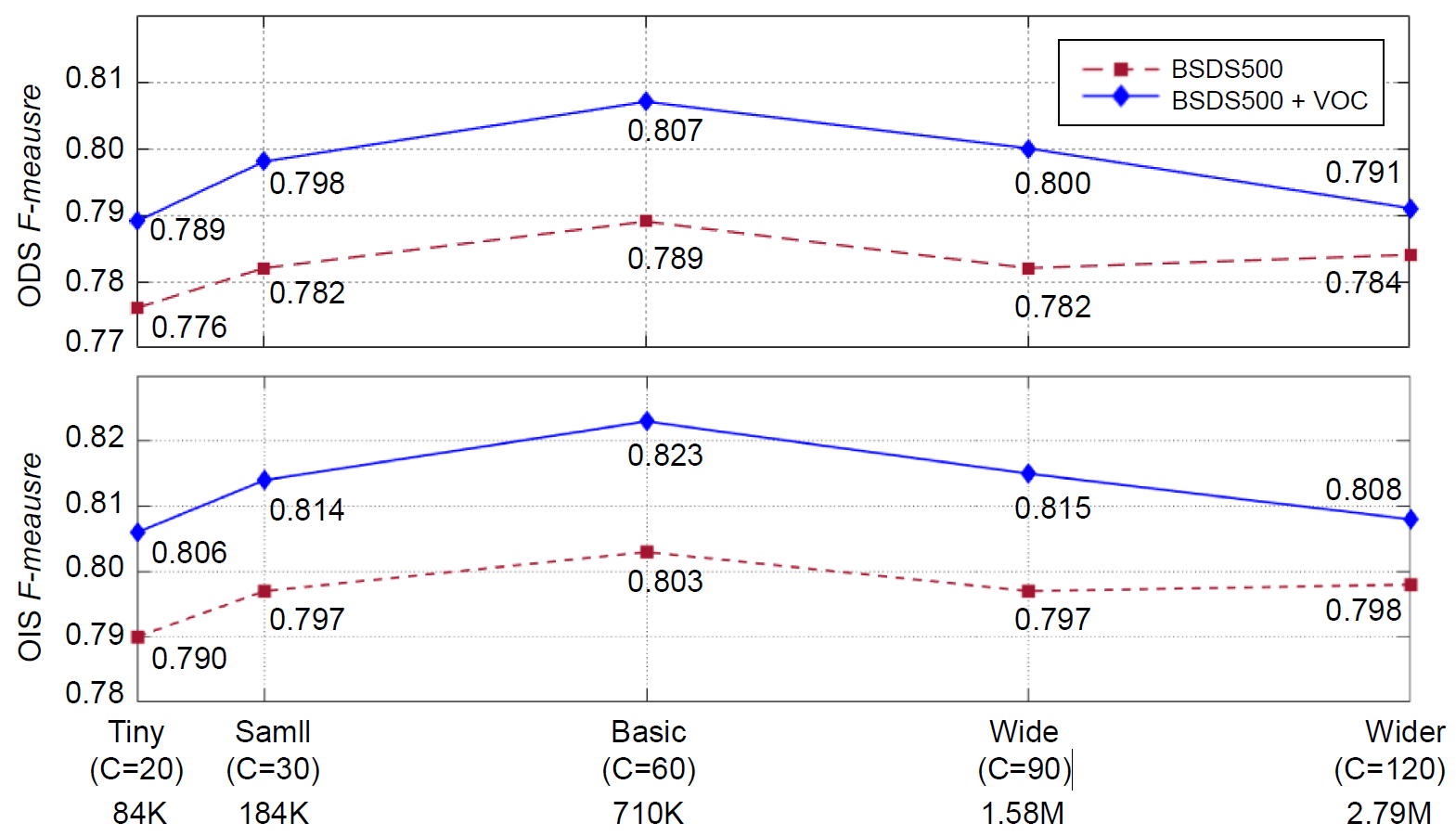}
    \caption{Exploration on the scalability of PiDiNet. The structure sizes are changed by slimming or widening the basic PiDiNet. Bottom row shows the number of parameters for each model. The models are trained with or without VOC dataset.}
    \label{fig:scalability}
\end{figure}

\begin{figure*}[t!]
    \centering
    \includegraphics[width=\linewidth]{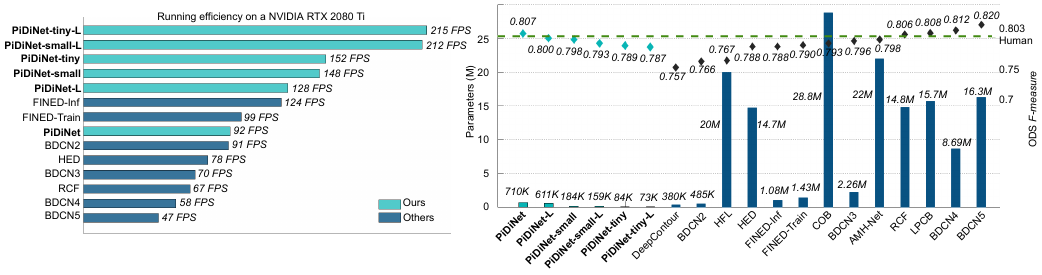}
    \caption{Comparison with other methods in terms of running efficiency (the inference speed as shown on the left part), memory storage (the number of parameters, shown on the right part) and detection performance (ODS \emph{F-measure} on BSDS500 dataset, shown on the right part). The running speeds of FINED~\cite{wibisono2020fined} are cited from the original paper, and the rest are evaluated by our implementations.}
    \label{fig:efficient}
\end{figure*}

\begin{table}[t!]
\captionsetup{labelfont={color=black},font={color=black}}
\caption{
Evaluation of model robustness under different image degradations~\cite{hendrycks2019imagenetc} on BSDS500 test set. There are 5 levels of intensity for each degradation type. We report the ODS/OIS ($\uparrow$) metrics under level 5 (L5) and level 1 (L1).}
\begin{center}
\setlength{\tabcolsep}{0.008\linewidth}
\resizebox*{\linewidth}{!}{
{\color{black}
\begin{tabular}{l|ccc|cccc}
\toprule
& \multicolumn{3}{c}{Noise} & \multicolumn{4}{c}{Blur} \\
Method & Gauss. & Shot & Impul. & Defoc. & Glass & Motion & Zoom \\
\midrule
Baseline (L5) & 0.014/0.014      & 0.041/0.041      & 0.010/0.010       &   0.528/0.540      & 0.724/0.744      & 0.580/0.595 & 0.552/0.567 \\
PiDiNet & 0.176/0.176      & 0.250/0.250      & 0.169/0.169       & 0.561/0.574       & 0.748/0.764      & 0.592/0.609       & 0.555/0.570 \\
\midrule
Baseline (L1) & 0.745/0.765       & 0.743/0.765     & 0.720/0.739      & 0.755/0.771      & 0.781/0.796      & 0.779/0/795       & 0.633/0.646 \\
PiDiNet & 0.734/0.750       & 0.739/0.756     & 0.707/0.722       & 0.769/0.786       & 0.796/0.811      & 0.790/0.805       & 0.637/0.653 \\
\midrule
& \multicolumn{3}{c}{Digital} & \multicolumn{4}{c}{Weather} \\
Method & Contr. & Pixel & JPEG & Snow & Frost & Fog & Brit. \\
\midrule
Baseline (L5) & 0.539/0.541      & 0.753/0.774      & 0.631/0.648 & 0.679/0.697      & 0.672/0.706       & 0.752/0.768    & 0.750/0.772 \\
PiDiNet & 0.657/0.678      & 0.768/0.785       & 0.648/0.661 & 0.676/0.690     & 0.712/0.722     & 0.753/0.772    & 0.763/0.782 \\
\midrule
Baseline (L1) & 0.777/0.795       & 0.795/0.813       & 0.773/0.786 & 0.763/0.780      & 0.776/0.794      &  0.787/ 0.803    & 0.795/0.813 \\
PiDiNet & 0.795/0.812       & 0.804/0.820      & 0.782/0.796 & 0.768/0.783      & 0.785/0.800      & 0.793/0.810    & 0.804/0.821 \\
\bottomrule
\end{tabular}
}
}
\end{center}
\label{tab:edge_noise}
\end{table}

\vspace{0.3em}
\noindent \textbf{CSAM, CDCM and Shortcuts.} We further conducted ablation experiments to validate the effectiveness of CSAM, CDCM, and residual structures. Specifically, we developed three network variants by removing these structures, as shown in \cref{table:moreablation}. 

On the one hand, the simple yet important addition of shortcuts increased ODS \emph{F-measure} by 5\%, as they can help preserve the gradient information captured by the previous layers. On the other hand, the attention mechanism in CSAM and dilation convolution in CDCM can give extra performance gains, while may also bring some computational cost. Therefore, they can be used to tradeoff between accuracy and efficiency. Specifically, in the following experiments, we kept both modules for better accuracy. Meanwhile, to pursue higher efficiency, we also built PiDiNet without CSAM and CDCM, which was denoted as PiDiNet-L (meaning a more lightweight version)

\vspace{0.3em}
\noindent \textbf{Network Scalability.} \quad
PiDiNet is highly compact with only 710K parameters and supports training from scratch with limited training data. Here, we explored the scalability of PiDiNet with different model complexities as shown in \cref{fig:scalability}. 
The models are denoted as PiDiNet-Tiny, PiDiNet-Small, PiDiNet-Wide, and PiDiNet-Wider, respectively.
For the fair comparison with other approaches, the models were trained twice using different training sets. First, the mixture of BSDS500 and VOC was used for training. Then, only BSDS500 was adopted as the training data.
Metrics were recorded on BSDS500 test set. 

Compared with the basic PiDiNet, smaller models have lower computational complexity while maintaining comparable performance in terms of both ODS and OIS scores. 
Moreover, training with more data consistently introduces notable accuracy gains. It should be noted that the basic PiDiNet can produce competitive ODS and OIS results as compared to HED~\cite{xie2017holistically} (\emph{i.e.}, 0.789 vs. 0.788 in ODS and 0.803 vs. 0.808 in OIS for PiDiNet vs. HED) even if it is trained without ImageNet pretraining. However, with limited training data, widening PiDiNet may cause the overfitting issue and results in degraded accuracy. 

\begin{table}[t!]
\caption{Comparison with other methods on BSDS500 dataset. $^\ddagger$ and $^\dagger$ indicate the GPU and CPU speeds with our implementations based on a NVIDIA RTX 2080 Ti GPU and Intel i7-8700 respectively.}
\begin{center}
\setlength{\tabcolsep}{0.02\linewidth}
\resizebox*{\linewidth}{!}{
\begin{tabular}{lccccc}
\toprule
Method & ImageNet & ODS & OIS & \multicolumn{2}{c}{FPS} \\
& PreTrain & & & GPU & \textcolor{black}{CPU} \\
& & & & & \\
\midrule
Human & - & .803 & .803 & - & \textcolor{black}{-} \\
\midrule
Canny~\cite{canny1986computational} & \xmark & .611 & .676 & - & \textcolor{black}{28} \\
Pb~\cite{martin2004pb} & \xmark & .672 & .695 & - & \textcolor{black}{-} \\
SCG~\cite{xiaofeng2012scg} & \xmark & .739 & .758 & - & \textcolor{black}{-} \\
SE~\cite{dollar2014se} & \xmark & .743 & .763 & - & \textcolor{black}{12.5} \\
OEF~\cite{hallman2015oef} & \xmark & .746 & .770 & - & \textcolor{black}{2/3} \\
\midrule
DeepContour~\cite{shen2015deepcontour} & \xmark & .757 & .776 & 1/30 & \textcolor{black}{-} \\
DeepEdge~\cite{bertasius2015deepedge} & \checkmark & .753 & .772 & 1/1000 & \textcolor{black}{-} \\
HFL~\cite{bertasius2015hfl} & \checkmark & .767 & .788 & 5/6 & \textcolor{black}{-} \\
CEDN~\cite{yang2016cedn} & \checkmark & .788 & .804 & 10 & \textcolor{black}{-} \\
HED~\cite{xie2017holistically} & \checkmark & .788 & .808 & 78$^\ddagger$ & \textcolor{black}{8$^\dagger$} \\
DeepBoundary~\cite{kokkinos2015deepboundary} & \checkmark & .789 & .811 & - & \textcolor{black}{-}\\
COB~\cite{maninis2016cob} & \checkmark & .793 & .820 & - & \textcolor{black}{-} \\
CED~\cite{wang2017ced} & \checkmark & .794 & .811 & - & \textcolor{black}{-} \\
AMH-Net~\cite{xu2018amhnet} & \checkmark & .798 & .829 & \textcolor{black}{-} \\
RCF~\cite{liu2019richer} & \checkmark & .806 & .823 & 67$^\ddagger$ & \textcolor{black}{10$^\dagger$} \\
LPCB~\cite{deng2018lpcb} & \checkmark & .808 & .824 & 30 & \textcolor{black}{-} \\
BDCN~\cite{he2019bidirectional} & \checkmark & .820 & .838 &  47$^\ddagger$ & \textcolor{black}{6$^\dagger$} \\
\textcolor{black}{DSCD\cite{deng2020dscd}} & \textcolor{black}{\checkmark} & \textcolor{black}{.813} & \textcolor{black}{.836} & \textcolor{black}{30} & \textcolor{black}{-} \\
\midrule
FINED-Inf~\cite{wibisono2020fined} & \xmark & .788 & .804 & 124 & \textcolor{black}{-} \\
FINED-Train~\cite{wibisono2020fined} & \xmark & .790 & .808 & 99 & \textcolor{black}{-} \\
\midrule
Baseline & \xmark & .798 & .816 & 96$^{\ddagger}$ & \textcolor{black}{11$^\dagger$} \\
PiDiNet & \xmark & .807 & .823 & 92$^{\ddagger}$ & \textcolor{black}{10$^\dagger$} \\
PiDiNet-L & \xmark & .800 & .815 & 128$^\ddagger$ & \textcolor{black}{14$^\dagger$} \\
PiDiNet-Small & \xmark & .798 & .814 & 148$^\ddagger$ & \textcolor{black}{21$^\dagger$} \\
PiDiNet-Small-L & \xmark & .793 & .809 & 212$^\ddagger$ & \textcolor{black}{28$^\dagger$} \\
PiDiNet-Tiny & \xmark & .789 & .806 & 152$^\ddagger$ & \textcolor{black}{31$^\dagger$} \\
PiDiNet-Tiny-L & \xmark & .787 & .804 & 215$^\ddagger$ & \textcolor{black}{51$^\dagger$} \\
\bottomrule
\end{tabular}
}
\end{center}
\label{table:bsds}
\end{table}

\begin{figure}[t!]
    \centering
    \includegraphics[width=1\linewidth]{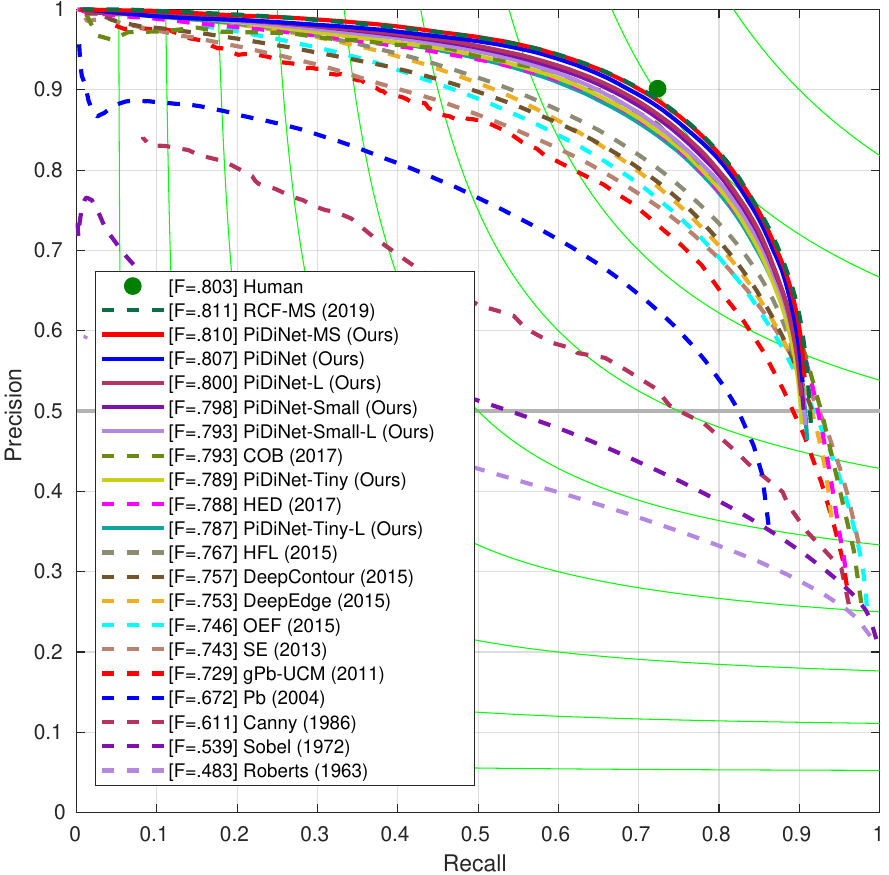}
    \caption{Precision-Recall curves of our models and some competitors on BSDS500 dataset.}
    \label{fig:bsds_pr}
\end{figure}

\begin{figure}[t!]
    \centering
    \includegraphics[width=1\linewidth]{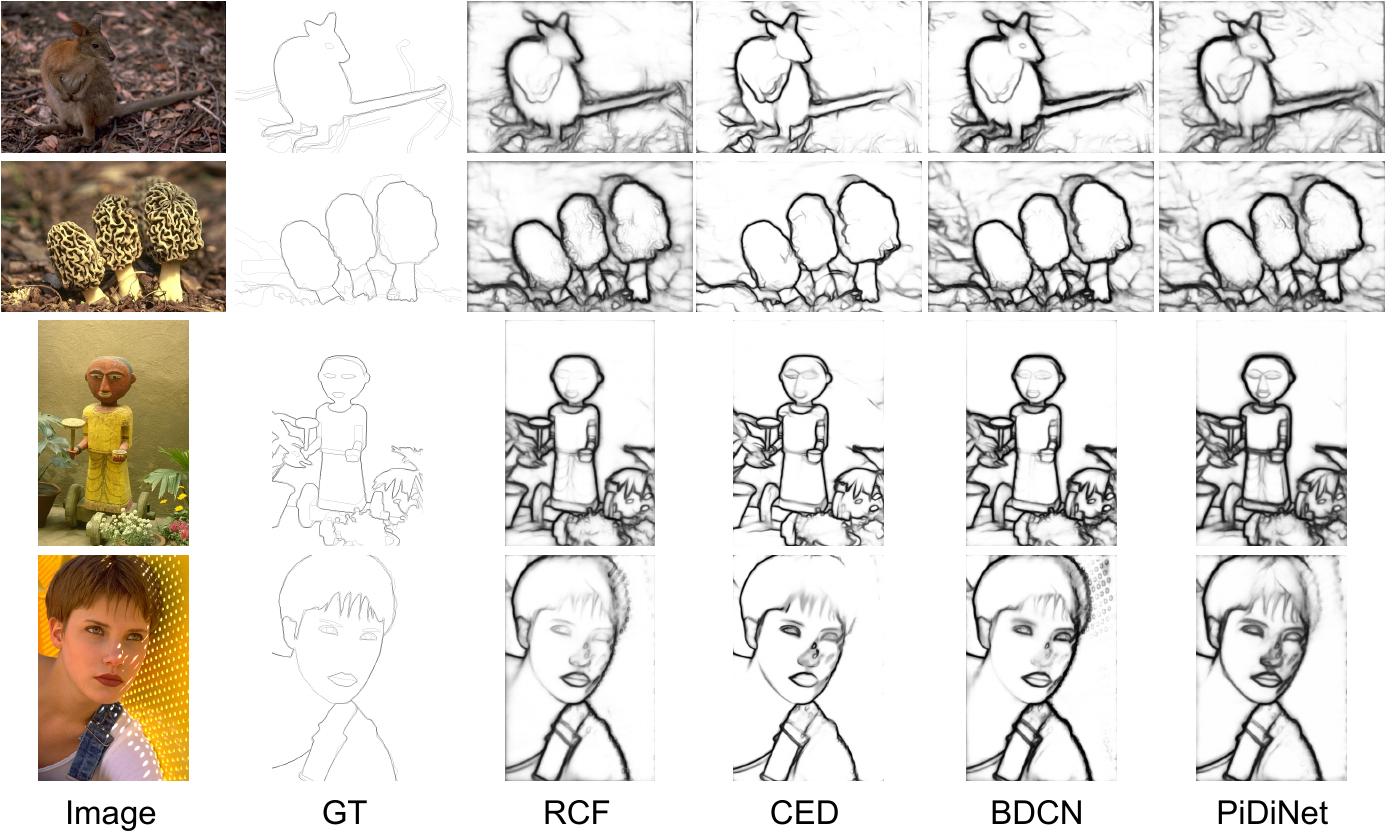}
    \caption{A qualitative comparison of network outputs with some other methods, including RCF~\cite{liu2019richer}, CED~\cite{wang2017ced} and BDCN~\cite{he2019bidirectional}. The first and second columns show the input and ground truth images, the rest columns show the corresponding output edge maps obtained by different methods.}
    \label{fig:bsds_visualization}
\end{figure}

\begin{table}[t!]
\caption{Comparison with other methods on NYUD dataset. $^\ddagger$ indicates the speeds with our implementations based on a NVIDIA RTX 2080 Ti GPU.}
\begin{center}
\setlength{\tabcolsep}{0.02\linewidth}
\resizebox*{\linewidth}{!}{
\begin{tabular}{l|c|c|c|c|c|c|c}
\toprule
Methods & ODS & OIS & ODS & OIS & ODS & OIS & FPS \\
\midrule 
gPb-UCM~\cite{arbelaez2010bsds} & .632 & .661 & & & & & 1/360 \\
gPb+NG~\cite{gupta2013gpbng} & .687 & .716 & & & & & 1/375 \\
SE~\cite{dollar2014se} & .695 & .708 & & & & & 5 \\
SE+NG+~\cite{gupta2014seng} & .710 & .723 & & & & & 1/15 \\
\midrule
\midrule
& \multicolumn{2}{c|}{RGB} & \multicolumn{2}{c|}{HHA} & \multicolumn{2}{c|}{RGB-HHA} & \\
\midrule
HED~\cite{xie2017holistically} & .720 & .734 & .682 & .695 & .746 & .761 & 62$^\ddagger$ \\
LPCB~\cite{deng2018lpcb} & .739 & .754 & .707 & .719 & .762 & .778 & - \\
RCF~\cite{liu2019richer} & .743 & .757 & .703 & .717 & .765 & .780 & 52$^\ddagger$ \\
AMH-Net~\cite{xu2018amhnet} & .744 & .758 & .716 & .729 & .771 & .786 & - \\
BDCN~\cite{he2019bidirectional} & .748 & .763 & .707 & .719 & .765 & .781 & 33$^\ddagger$ \\
\midrule
PiDiNet & .733 & .747 & .715 & .728 & . 756 & .773 & 62$^\ddagger$ \\
PiDiNet-L & .728 & .741 & .709 & .722 & .754 & .770 & 88$^\ddagger$ \\
PiDiNet-Small & .726 & .741 & .705 & .719 & .750 & .767 & 115$^\ddagger$ \\
PiDiNet-Small-L & .721 & .736 & .701 & .713 & .746 & .763 & 165$^\ddagger$ \\
PiDiNet-Tiny & .721 & .736 & .700 & .714 & .745 & .763 & 140$^\ddagger$ \\
PiDiNet-Tiny-L & .714 & .729 & .693 & .706 & .741 & .759 & 206$^\ddagger$ \\
\bottomrule
\end{tabular}
}
\end{center}
\label{table:nyud}
\end{table}

\vspace{0.3em}
\noindent \textcolor{black}{\textbf{Model Robustness.} \quad\
Following~\cite{ovadia2019trustmodels},
we evaluated the robustness of PiDiNet to image degradations like noises, corruptions, and perturbations. For each image in the test set of BSDS500, a degraded image was first synthesized following~\cite{hendrycks2019imagenetc} and then fed to the  PiDiNet and baseline (which were trained on clean data under the setting described in \cref{sec:edge_sota}), respectively. We report the ODS/OID metrics to make the comparison. It can be seen in \cref{tab:edge_noise} that PiDiNet outperforms the baseline in most degradation cases, demonstrating the robustness of our PDC against vanilla convolution on edge detection. 
}

\subsubsection{Comparison with the State-of-the-Art Methods}
\label{sec:edge_sota}

\noindent \textbf{On BSDS500 dataset.} \quad Following previous methods~\cite{liu2019richer,deng2018lpcb,he2019bidirectional}, we used the mixture of BSDS500 and VOC as the training set for fair comparison.
Metrics were recorded on the test set. We compared our methods with prior edge detection approaches including traditional ones \cite{canny1986computational,martin2004pb,xiaofeng2012scg} and recent CNN based ones \cite{bertasius2015hfl,yang2016cedn,xie2017holistically,xu2018amhnet,liu2019richer,deng2018lpcb,he2019bidirectional}. Results are presented in \cref{table:bsds} and \cref{fig:bsds_pr}. 

It can be observed that our baseline model achieves comparable results, \emph{i.e.}, with ODS of 0.798 and OIS of 0.816, already beating most CNN based models like CED~\cite{wang2017ced}, DeepBoundary~\cite{kokkinos2015deepboundary}, and HED~\cite{xie2017holistically}. With our PDC, PiDiNet can further boost the performance with ODS of 0.807, being the same level as RCF~\cite{liu2019richer} while still achieving nearly 100 FPS on the GPU \textcolor{black}{(it should be noted that PiDiNet is slightly slower than the baseline as the RPDC is converted to $5\times 5$ vanilla convolutions)}. The fastest version PiDiNet-Tiny-L, can also achieve comparable prediction performance with more than 200 FPS on GPU and 50 FPS on CPU, further demonstrating the effectiveness of our methods. Furthermore, all of our modes were trained from scratch
without the ImageNet pretraining. A more detailed comparison in terms of network complexity, running efficiency, and accuracy can be seen in \cref{fig:efficient}. We also illustrated some qualitative results in Figure~\ref{fig:bsds_visualization}, from which we can see our method obtained high-quality edge maps that were comparable to those by the state-of-the-art approaches. 

\vspace{0.3em}
\noindent \textbf{On NYUD dataset.} \quad We utilized
the training and validation sets of the NYUD dataset and augmented them with flipping
(2×), scaling (3×), and rotation (4×) to compose our
training data. The comparison results on the NYUD test set are illustrated in \cref{table:nyud}
Following prior works, `RGB-HHA' results are obtained by averaging the output edge maps from RGB image and HHA image to get the final edge map. 

\begin{table}[t!]
\caption{Comparison with other methods on Multicue dataset. $^\ddagger$ indicates the speeds with our implementations based on a NVIDIA RTX 2080 Ti GPU.}
\begin{center}
\setlength{\tabcolsep}{0.015\linewidth}
\resizebox*{\linewidth}{!}{
\begin{tabular}{l|c|c|c|c|c}
\toprule
Method & \multicolumn{2}{c|}{Boundary} & \multicolumn{2}{c|}{Edge} & FPS \\
\cmidrule(r){2-5}
 & ODS & OIS & ODS & OIS & \\
\midrule 
Human~\cite{mely2016multicue} & .760 (.017) & - & .750 (.024) & - & - \\
\midrule
Multicue~\cite{mely2016multicue} & .720 (.014) & - & .830 (.002) & - & - \\
HED~\cite{xie2017holistically} & .814 (.011) & .822 (.008) & .851 (.014) & .864 (.011) & 18$^\ddagger$ \\
RCF~\cite{liu2019richer} & .817 (.004) & .825 (.005) & .857 (.004) & .862 (.004) & 15$^\ddagger$ \\
BDCN~\cite{he2019bidirectional} & .836 (.001) & .846 (.003) & .891 (.001) & .898 (.002) & 9$^\ddagger$ \\
\textcolor{black}{DSCD~\cite{deng2020dscd}} & \textcolor{black}{.828 (.003)} & \textcolor{black}{.835 (.004)} & \textcolor{black}{.871 (.007)} & \textcolor{black}{.876 (.002)} & \textcolor{black}{-} \\
\midrule
PiDiNet & .818 (.003) & .830 (.005) & .855 (.007) & .860 (.005) & 17$^\ddagger$ \\
PiDiNet-L & .810 (.005) & .822 (.002) & .854 (.007) & .860 (.004) & 23$^\ddagger$ \\
PiDiNet-Small & .812 (.004) & .825 (.004) & .858 (.007) & .863 (.004) & 31$^\ddagger$ \\
PiDiNet-Small-L & .805 (.007) & .818 (.002) & .854 (.007) & .860 (.004) & 44$^\ddagger$ \\
PiDiNet-Tiny & .807 (.007) & .819 (.004) & .856 (.006) & .862 (.003) & 43$^\ddagger$ \\
PiDiNet-Tiny-L & .798 (.007) & .811 (.005) & .854 (.008) & .861 (.004) & 56$^\ddagger$ \\
\bottomrule
\end{tabular}
}
\end{center}
\label{table:multicue}
\end{table}

From \cref{table:nyud} we can see that our PiDiNets produce competitive results among the state-of-the-art methods while being highly efficient.

\vspace{0.3em}
\noindent \textbf{On Multicue dataset.} \quad Following~\cite{liu2019richer,he2019bidirectional}, we randomly split the Multicue dataset to training and evaluation sets with a ratio of 8:2. This process was independently repeated twice more. The metrics were then recorded from the three runs. During training, we also augmented each training image with flipping (2$\times$), scaling (3$\times$), and rotation ($16\times$), then randomly cropped them with size 500$\times$500. The comparison results with other methods are shown in \cref{table:multicue}. As we can see, our PiDiNets achieve promising results with high efficiency. For example, our PiDiNet produces higher ODS and OIS scores than HED with comparable FPS. In addition, our PiDiNet-Tiny achieves comparable performance to HED with over $2\times$ speedup. This further demonstrates the effectiveness of our PiDiNet in terms of both accuracy and efficiency.

\subsection{Bi-PiDiNet on Object Recognition}
\label{sec:experiments-recognition}

\noindent \textbf{Datasets.} \quad We conducted our experiments on CIFAR-100~\cite{krizhevsky2009cifar}, ImageNet~\cite{deng2009imagenet}, and five facial datasets for object recognition. The CIFAR-100 dataset is comprised of 5K training images and 1K testing images with 100 classes. The ImageNet dataset with 1K classes (also termed as ILSVRC2012)~\cite{ILSVRC15} contains 1.2 million images for training and 50K mages for validation. In addition, we employed five facial datasets for training and evaluation, including  CASIA-WebFace~\cite{yi2014webface}, LFW~\cite{huang2008lfw}, CPLFW~\cite{zheng2018crosspose}, CALFW~\cite{zheng2017crossage}, and YTF~\cite{wolf2011youtubeface}. The CASIA-WebFace dataset was further washed up by removing images with wrong labels, leaving 0.45 million images.

\vspace{0.3em}
\noindent \textbf{Evaluation Metrics.} \quad The top-1 and top-5 accuracies were used for evaluation on CIFAR-100 and ImageNet. For facial recognition, the overall testing accuracy and the area under the curve (AUC) metrics were adopted as the metrics. To evaluate the network complexity regarding computational cost and memory storage, the number of BOPs, FLOPs, the total OPs (\#OPs = \#FLOPs + $1/64\times$\#BOPs), and the memory consumption were used following~\cite{liu2020reactnet,zhao2022bonn}. We excluded the last linear layer when calculating the complexity of network since its computational cost is negligible (\eg only 0.3\% for ResNet-18 backbone).

\vspace{0.3em}
\noindent \textbf{Training settings.} \quad
Instead of adopting the advanced multi-stage training scheme~\cite{liu2020reactnet,martinez2020realtobinary}, all our models were trained from scratch with a moderate number of epochs (60 epochs for image classification and 20 epochs for facial recognition). Meanwhile, we avoided the use of any gradient approximation techniques~\cite{xu2021frequencydomain,liu2018birealnet} but the original STE in the same way as in~\cite{bengio2013ste} for gradient back-propagation. Our implementations were based on the Pytorch 1.7 library~\cite{paszke2019pytorch} with Adam optimizer~\cite{kingma2014adam}. Dataset-specific implementation details will be given in the following individual sections.

\subsubsection{Model Analysis}
\label{sec:ablation-image}

\begin{figure}[t!]
    \centering
    %\vspace{-0.1in}
    \includegraphics[width=\linewidth]{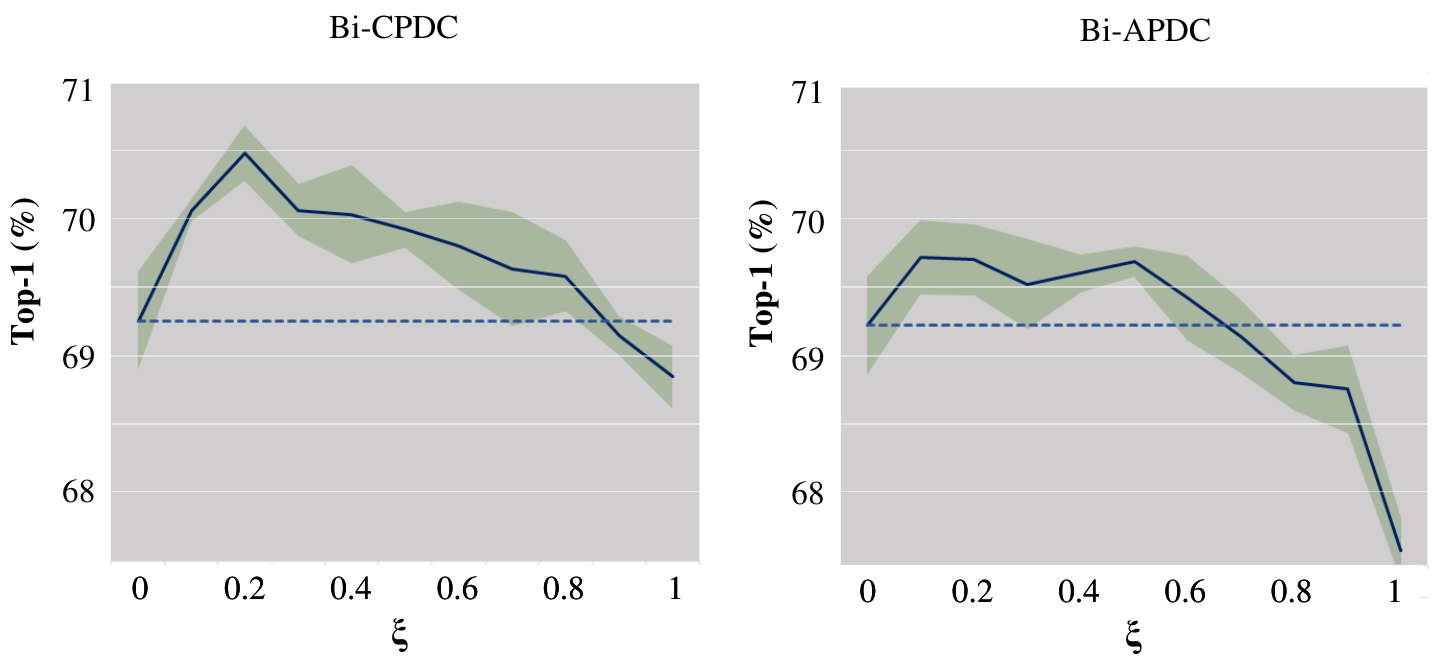}
    \caption{
    Influences of $\xi$ (from 0 to 1 with step 0.1) on the Top-1 accuracy (\%) on CIFAR-100 datasets, for Bi-CPDC and Bi-APDC based models respectively. The solid line and shadow record the average value and the standard deviation of accuracies in each $\xi$ based on five independent runs, the dashed line represents the average accuracy of the baseline model.
    }
    \label{fig:cifar100}
\end{figure}

The ablation study was conducted for both natural image classification and facial recognition. For the former task, ResNet-18 was used as the baseline model. For the latter task, the 20-layer CNN architecture in~\cite{liu2017sphereface} was adopted.
For training on CIFAR-100, random padding, random crop (to size $32 \times 32$), and random horizontal flipping were used for data augmentation. The initial learning rate was set to 0.001 and reduced by 0.1 at epoch 45 and 55. Batch size was set to 64. For evaluation, we trained each model for five times independently and calculated the mean values and standard deviations of their quantitative results. For training on ImageNet, the input images were augmented with random cropping (to size $224 \times 224$) and random horizontal flipping. The setting of learning rate and epochs was the same as for CIFAR-100, and batch size was set to 256. For the facial dataset, following~\cite{liu2017sphereface}, we trained the models on CASIA-WebFace, and evaluated them on the challenging CALFW, CPLFW, and YTF datasets. During training, each image was aligned with the five facial landmarks detected by MTCNN~\cite{zhang2016joint} and cropped to size $112 \times 96$. The initial learning rate was 0.001, which was decayed by 0.1 at epoch 12 and 17. Batch size was 128. During testing, the test images were cropped to the same size as used during training. The face representation was then obtained by concatenating the features from the original and horizontally flipped images. If not specified, we only built our BCNN models without further improving the architectures for simplicity in the ablation study (see \cref{fig:bipidinet} (a)).

\begin{table}[t!]
\caption{Ablation study on LBP. case 1: the binary weight kernels were randomly initialized and fixed; case 2: same as case 1, with the condition that ``uniform'' patterns constitute the major part (\ie 80\%) of kernels. W/A means the number of bits used in weights and activations respectively.}
\label{tab:ablation-lbp}
\begin{center}
\setlength{\tabcolsep}{0.008\linewidth}
\resizebox*{\linewidth}{!}{
\begin{tabular}{lccc}
\toprule
Model & W/A & Top-1 (\%) & Top-5 (\%) \\
\midrule
Full-precision & 32/32 & 75.61 & 93.05 \\
\midrule
Baseline (vanilla BCNN) & 1/1 & 69.26 $\pm$ 0.36 & 88.92 $\pm$ 0.21 \\
Fixed LBP patterns - case 1 & 1/1 & 68.80 $\pm$ 0.11 & 88.90 $\pm$ 0.16 \\
Fixed LBP patterns - case 2 & 1/1 & 68.91 $\pm$ 0.18 & 88.94 $\pm$ 0.26 \\
Learnable LBP patterns (our method) & 1/1 & 70.49 $\pm$ 0.20 & 89.57 $\pm$ 0.20 \\
\bottomrule
\end{tabular}
}
\end{center}
\end{table}

\vspace{0.3em}
\noindent \textbf{Selection of $\xi$.} \quad $\xi$ is a key hyper-parameter that determines ratio of input channels fed into our Bi-PiDiNet. To study its effect on the performance, we constructed different models with different values of $\xi$. Then, we trained these models on CIFAR-100 with the same setting as mentioned above. The results are shown in \cref{fig:cifar100}.

As we can see, compared to the baseline with $\xi=0$, the performance was gradually improved as $\xi$ was increased to around 0.2. This demonstrates that our Bi-PiDiNet can provide complementary information to vanilla binary convolutions to produce performance gain. However, the models suffer from notable performance drop if $\xi$ was further increased. Since models with $\xi=0.2$ achieves the best performance, it was used as the default setting in our network. It should be noted that we did not observe accuracy gain from Bi-RPDC on CIFAR-100 due to the small resolutions of this dataset. The large padding size in Bi-RPDC (\ie 2) may degrade the performance.

\begin{figure}[t!]
    \centering
    \includegraphics[width=\linewidth]{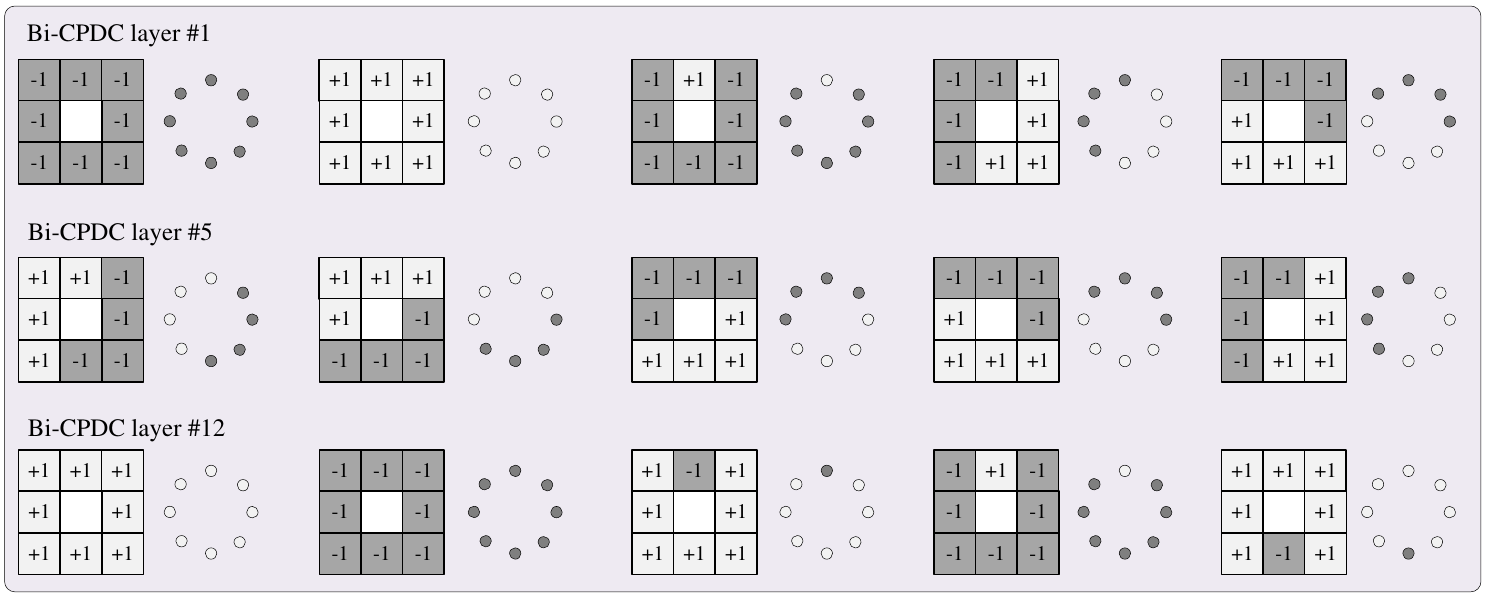}
    \caption{The top-5 most frequent weight kernels learned in certain layers in our Bi-CPDC based model (trained on ImageNet dataset) with the associated LBP patterns. It is shown that Bi-CPDC tends to capture regular patterns, \emph{i.e.}, the ``uniform'' patterns in~\cite{ojala2002multiresolution}, which correspond to common micro-structures in natural images such as spots, flat areas, edges, \emph{etc.}}
    \label{fig:extract}
\end{figure}

\begin{figure}[t!]
    \centering
    \includegraphics[width=\linewidth]{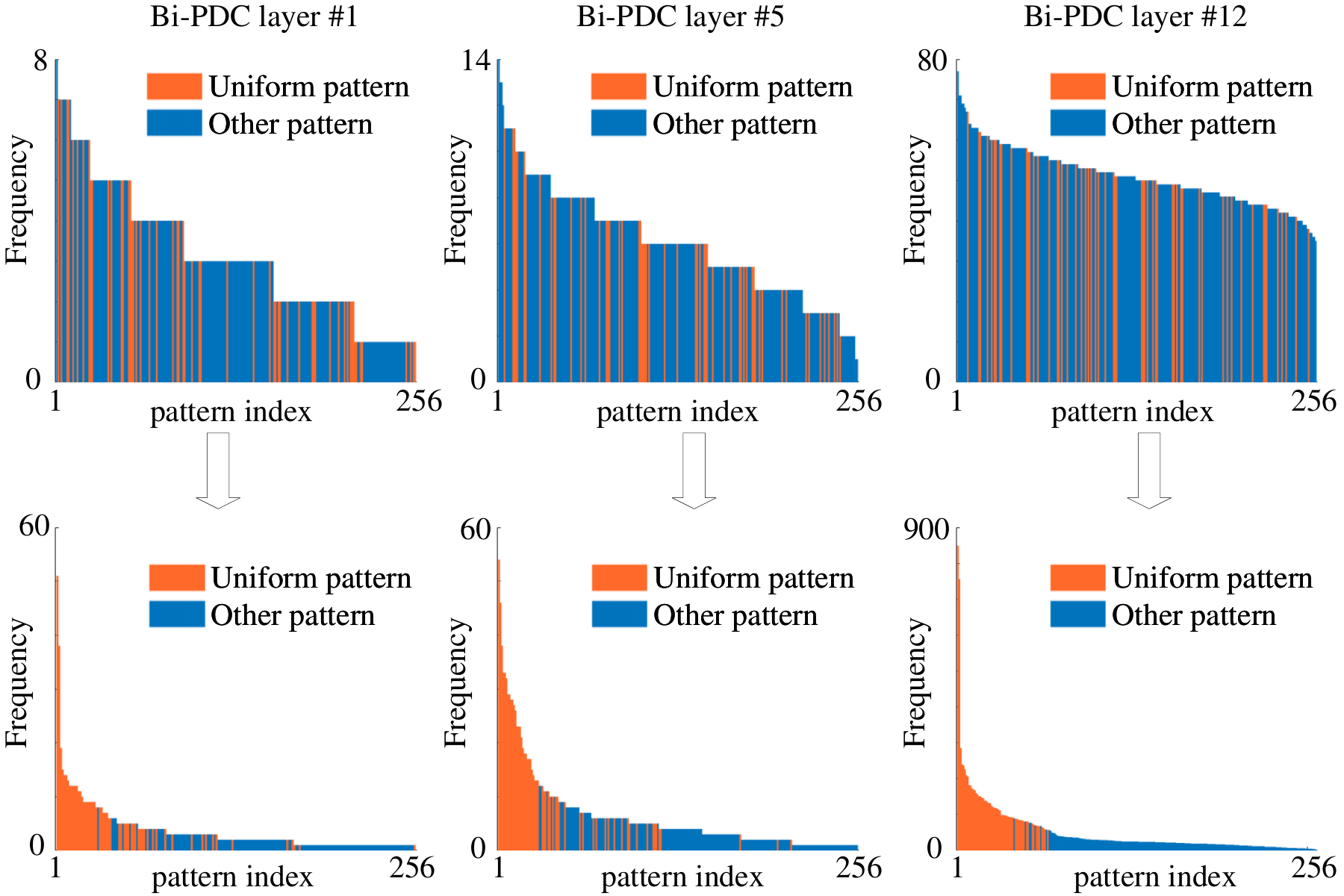}
    \caption{Top row: statistics of the $2^p(=256)$ patterns before training. Bottom row: statistics of the patterns after training. The patterns were sorted according to the frequencies. The model was trained on ImageNet.}
    \label{fig:lbp_code}
\end{figure}

\begin{table}[t!]
\caption{Model selection on ImageNet and facial datasets with the corresponding accuracy values (\%). The bold and underline values denote the best and the second best ones in each column, respectively. W/A means the number of bits used in weights and activations respectively. }
\label{tab:ablation-selection-imageNet}
\begin{center}
\setlength{\tabcolsep}{0.02\linewidth}
\resizebox*{\linewidth}{!}{
\begin{tabular}{lccccc}
\toprule
Model & W/A & ImageNet & CALFW & CPLFW & YTF \\
\midrule
Full-precision & 32/32 & 70.3 & 90.78 & 76.18 & 93.20 \\
\midrule
Vanilla BConv & 1/1 & 60.3 & 86.13 & 70.85 & 92.10 \\
\textbf{Bi-CPDC} & 1/1 & \textbf{61.6} & \underline{86.62} & \underline{73.30} & 92.68\\
Bi-APDC & 1/1 & \underline{60.9} & \textbf{87.03} & \textbf{73.47} & \underline{92.88} \\
Bi-RPDC & 1/1 & 60.7 & 86.42 & 71.82 & \textbf{92.96} \\
\bottomrule
\end{tabular}
}
\end{center}
\end{table}

\begin{table*}[t!]
\caption{Ablation study on our architectural improvement (On ImageNet).}
\label{tab:ablation-architecture-improve}
\begin{center}
\setlength{\tabcolsep}{0.01\linewidth}
\resizebox*{\linewidth}{!}{
\begin{tabular}{lcc|c|cc|c|cc}
\toprule
Method & \begin{tabular}{@{}c@{}}FLOPs\\($\times 10^8$)\end{tabular} & \begin{tabular}{@{}c@{}}BOPs\\($\times 10^8$)\end{tabular} & \begin{tabular}{@{}c@{}}OPs\\($\times 10^8$)\end{tabular} & \begin{tabular}{@{}c@{}}FP-params\\($\times 10^6$)\end{tabular} & \begin{tabular}{@{}c@{}}B-params\\($\times 10^6$)\end{tabular} & \begin{tabular}{@{}c@{}}Total memory\\(M bits)\end{tabular} & \begin{tabular}{@{}c@{}}Top-1\\(\%)\end{tabular} & \begin{tabular}{@{}c@{}}Top-5\\(\%)\end{tabular} \\
\midrule
Full-precision & 17.70 & 0 & 17.70 & 11.18 & 0 & 358 & 70.3 & 89.5 \\
\midrule
Architecture with vanilla BConv before improving & 1.45 & 16.76 & 1.71 & 0.20 & 10.99 & 17 & 60.3 & 82.2 \\
Architecture with Bi-PDC before improving & 1.46 & 16.40 & 1.72 & 0.21 & 10.74 & 17 & 61.6 & 82.9 \\
\textbf{Architecture with Bi-PDC after improving (Bi-PiDiNet)} & 0.25 & 42.74 & 0.92 & 0.04 & 19.23 & 20 & 62.8 & 83.8 \\

\bottomrule
\end{tabular}
}
\end{center}
\end{table*}

\begin{table*}[t!]
\captionsetup{labelfont={color=black},font={color=black}}
\centering
\caption{Evaluation of model robustness under different degradations on CIFAR100-C. The average accuracy (\%) is based on five independent runs. For each run, the model was trained on the original CIFAR-100 dataset and tested on CIFAR100-C.}
\label{Tab:noise_cifar}
\setlength{\tabcolsep}{0.01\linewidth}
\resizebox{\linewidth}{!}{
{\color{black}
\begin{tabular}{l|ccc|cccc|cccc|cccc}
\toprule
                                     & \multicolumn{3}{c}{Noise}                                                           & \multicolumn{4}{c}{Blur}                                                                                         & \multicolumn{4}{c}{Weather}                                                                                 & \multicolumn{4}{c}{Digital}                                                                                       \\
Method         & Gauss. & Shot & Impul. & Defoc. & Glass & Motion & Zoom  & Snow & Frost & Fog & Brit. & Contr. & Elastic & Pixel & JPEG  \\ 
\midrule
Baseline (level 5)    & 5.15                       &  5.45                    & 4.02   & 3.31                      & 5.36                      & 3.68                     & 4.18 & 11.04                    &  6.82                    &   3.28                & 18.57 &  1.12                      & 7.17                      &  5.24                    & 9.41 \\
APDC           & 5.15                      & 5.23                    & 4.95  & 5.82                     &  3.97                   & 6.27                      & 6.41 &  10.80                  &  6.39                   &  4.12                  & 22.6 &  3.00                    &  7.29                      &  4.48                    & 10.02 \\
CPDC           & 4.69                      & 4.92                   &  4.91  &  7.08                    & 4.50                    & 7.82                      & 8.06 & 11.44                   &  6.94                    & 5.22                   & 24.44 & 4.32                      &  8.68                      &  3.35                    & 10.88 \\ 
\midrule
Baseline (level 1)    & 14.94                       &  16.49                    & 18.45   & 17.35                      & 7.47                      & 10.70                     & 8.83 & 17.24                    &  17.22                   &   14.16                & 20.71 &  12.08                     & 11.53                     &  17.51                    & 14.21 \\
APDC           &13.42                       & 15.55                    & 20.71  & 22.35                     & 4.96                    & 14.43                      & 12.33 &  20.20                   & 18.78                    & 18.49                   & 25.94 &  16.60                      &  14.90                      & 21.55                     & 16.23 \\
CPDC           & 12.33                       & 15.18                   & 22.65  & 25.15                       & 5.59                      & 17.03                      & 14.65 & 21.95                   &  20.07                   &21.06                   & 28.07 &  18.98                    & 17.18                       & 22.07                     & 17.31 \\ 
\bottomrule
\end{tabular}
}
}
\end{table*}

\vspace{0.3em}
\noindent \textbf{Learnable vs. Fixed LBP Descriptors.} \quad Bi-PDC can be regarded as learnable LBP descriptors. Specifically, in traditional LBP methods, the $2^p$ possible codes were manually divided into several groups representing certain micro-structures, \emph{e.g.}, the $p+1$ ``uniform'' patterns in~\cite{ojala2002multiresolution}, or the dominant patterns in~\cite{liao2009dominant}. 
In contrast, Bi-PDC automatically learns which patterns are important during training. To validate the effectiveness of the learnable LBP patterns in our Bi-PDC, we developed two network variants by fixing the weight kernels of Bi-PDC. From the quantitative results in \cref{tab:ablation-lbp}, we can see that the models with fixed LBP patterns suffer notable accuracy drop. In contrast, learnable LBP patterns in our Bi-PDC facilitates our network to achieve much higher performance.

As the convolution operation essentially calculates the cosine similarities between the extracted pixel patterns and the weight kernels,\footnote{More strictly, the convolution process calculates the inner products between pixel sequences and weight kernels. While in binary convolution where there are only \{-1, +1\} numbers, the inner products turn to cosine similarities with a certain scaling factor.} only those pattern codes with high similarity with the kernels give high responses in the output. Therefore, we further visualize the learned weight kernels to investigate the learned patterns. 
For example, as shown in \cref{fig:extract} and \cref{fig:lbp_code} for Bi-CPDC, the patterns in the kernel are prone to associate with ``uniform'' patterns in LBP that usually have more physical meanings~\cite{ojala2002multiresolution}.

\vspace{0.3em}
\noindent \textbf{Selection of Bi-PDC.} \quad
We conduct ablation experiments to investigate different types of Bi-PDC. Comparative results are presented in
\cref{tab:ablation-selection-imageNet}. As we can see, Bi-CPDC achieves the best performance on ImageNet while Bi-APDC produces the highest accuracy on facial datasets.
Therefore, we adopted Bi-CPDC and Bi-APDC as the default settings for ImageNet and facial datasets. 

\begin{figure}[t!]
    \centering
    \includegraphics[width=\linewidth]{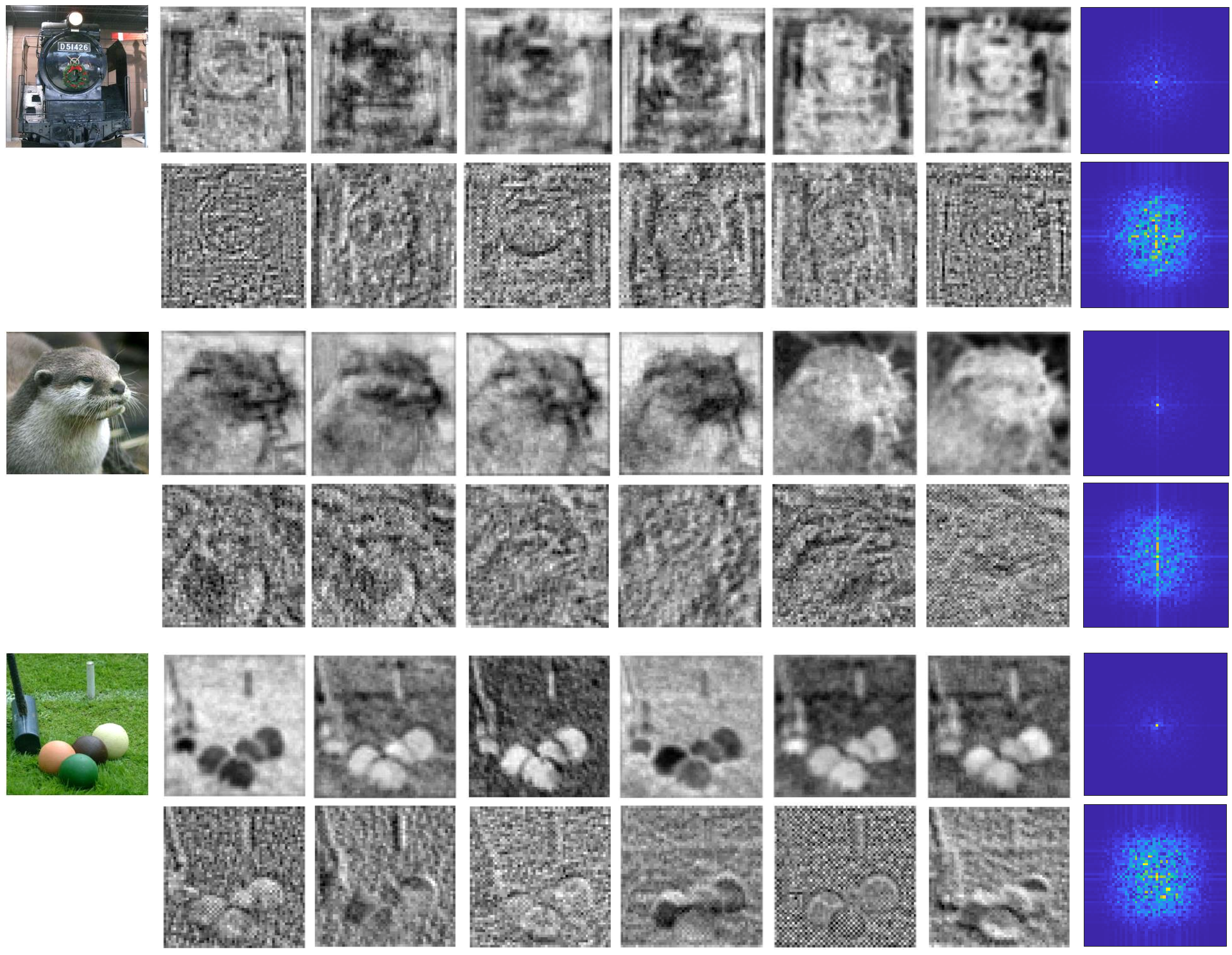}
    \caption{
    The first row for each input image: feature maps generated by vanilla BConv and the corresponding FFT2 map (by firstly averaging all the feature maps, then doing FFT2 transformation). The second row: feature maps generated by Bi-PDC (Bi-CPDC in this visualization) and the FFT2 map.
    }
    \label{fig:features_imagenet_compare}
\end{figure}

\begin{table}[t!]
\caption{Comparison of recognition performances on ImageNet dataset with the state-of-the-art methods. W/A means the number of bits used in weights and activations respectively. ReActNet$^*$ indicates the model trained with the RSign and RPReLU functions in~\cite{liu2020reactnet}. FP means full-precision.}
\label{tab:comparison-imagenet}
\begin{center}
\setlength{\tabcolsep}{0.010\linewidth}
\resizebox*{\linewidth}{!}{
\begin{tabular}{lccc|cc}
\toprule
Method & W/A & \multicolumn{2}{c}{ResNet-18} & \multicolumn{2}{c}{ResNet-34}\\
\cmidrule(r){3-6}
& & Top-1 (\%) & Top-5 (\%) & Top-1 (\%) & Top-5 (\%) \\
\midrule
\textcolor{black}{FP-vanilla Conv} & \textcolor{black}{32/32} & \textcolor{black}{70.3} & \textcolor{black}{89.5} & \textcolor{black}{73.5} & \textcolor{black}{91.3} \\
\textcolor{black}{FP-CPDC} & \textcolor{black}{32/32} & \textcolor{black}{70.4} & \textcolor{black}{89.7} & \textcolor{black}{73.9} & \textcolor{black}{91.7} \\
\midrule
ABC-Net~\cite{lin2017abcnet} & 1/1 & 42.7 & 67.6 & 52.4 & 76.5 \\
XNOR-Net~\cite{rastegari2016xnor} & 1/1 & 51.2 & 73.2 & - & - \\
DoReFa-Net~\cite{zhou2016dorefa} & 1/2 & 53.4 & - & - & - \\
Bi-Real~\cite{liu2018birealnet} & 1/1 & 56.4 & 79.5 & 62.2 & 83.9 \\
XNOR++~\cite{bulat2019xnorpp} & 1/1 & 57.1 & 79.9 & - & - \\
IR-Net~\cite{qin2020forward} & 1/1 & 58.1 & 80.0 & 62.9 & 84.1 \\
BONN~\cite{zhao2022bonn} & 1/1 & 59.3 & 81.6 & - & - \\
Han~\etal~\cite{han2020noisysup} & 1/1 & 59.4 & 81.7 & - & - \\
RBNN~\cite{Mingbao2020rbnn} & 1/1 & 59.9 & 81.9 & 63.1 & 84.4 \\
ReActNet$^*$~\cite{liu2020reactnet} & 1/1 & 60.2 & 82.0 & 63.8 & 84.9 \\
FDA-BNN~\cite{xu2021frequencydomain} & 1/1 & 60.2 & 82.3 & - & - \\
Bi-half~\cite{li2022equalbits} & 1/1 & 60.4 & 82.9 & 64.2 & 85.4 \\
ReCU~\cite{Xu_2021_recu} & 1/1 & 61.0 & 82.6 & 65.1 & 85.8 \\
\midrule
\emph{Bi-PiDiNet} & 1/1 & \textbf{62.8} & \textbf{83.8} & \textbf{66.3} & \textbf{86.5} \\
\bottomrule
\end{tabular}
}
\end{center}
\end{table}

\vspace{0.3em}
\noindent \textbf{Architectural improvement.} \quad
To demonstrate the effectiveness of our method on architectural improvement for BCNNs, we compared the architectures with and without our improvement in terms of model complexity and accuracy, as shown in \cref{tab:ablation-architecture-improve}. In detail, the original ResNet backbone contains 64, 128, 256, 512 channels in each stage respectively. We widened the network by adopting 128, 192, 384, 768 channels accordingly. It can be seen further improving the BCNN architecture (see \cref{fig:bipidinet} (b)) leads to a nearly $2\times$ reduction on the computational cost (\#OPs) and 1.2\% Top-1 gain on the ImageNet dataset, with only a minor increase in memory storage.

\vspace{0.3em}
\noindent \textbf{High-frequency information extraction.} \quad
Our Bi-PDC can extract complementary high-frequency information to improve the performance of BCNNs. To validate this, we visualize the intermediate feature maps generated by vanilla BConv and Bi-PDC as well as the corresponding frequency maps (the FFT2 results of the averaged feature map) on ImageNet. 
As shown in \cref{fig:features_imagenet_compare}, the feature maps produced by our Bi-PDC can capture more high-frequency information as compared to vanilla BConv. From the perspective of frequency domain, vanilla BConv focuses on the low-frequency information, while our Bi-PDC pays more attention to high-frequency details, which facilitates our network to produce higher accuracy. 

\begin{table*}[t!]
\caption{Detailed comparison on network complexities based on the ResNet-18 backbone. Additional training was composed of multi-stage training and knowledge distillation following~\cite{liu2020reactnet}. Specifically, on the first training stage, we only binarized activations while kept model weights full-precision. On the second stage, we inherited the full-precision weights from the previous stage and binarize both weights and activations. For both stages, the full-precision ResNet-34 network was used as the teacher network.}
\label{tab:resnet18-imagenet}
\begin{center}
\setlength{\tabcolsep}{0.017\linewidth}
\resizebox*{\linewidth}{!}{
\begin{tabular}{lcc|c|cc|c|cc}
\toprule
Method &  \multicolumn{3}{l}{Computations} & \multicolumn{3}{l}{Parameters} & \begin{tabular}{@{}c@{}}Additional\\training\end{tabular} & \begin{tabular}{@{}c@{}}Top-1\\(\%)\end{tabular} \\
\cmidrule(r){2-7}
& \begin{tabular}{@{}c@{}}FLOPs\\($\times 10^8$)\end{tabular} & \begin{tabular}{@{}c@{}}BOPs\\($\times 10^8$)\end{tabular} & \begin{tabular}{@{}c@{}}OPs\\($\times 10^8$)\end{tabular} &
\begin{tabular}{@{}c@{}}FP-params\\($\times 10^6$)\end{tabular} & \begin{tabular}{@{}c@{}}B-params\\($\times 10^6$)\end{tabular} & \begin{tabular}{@{}c@{}}Total memory\\(M bits)\end{tabular} & & \\
\midrule
Full-precision & 17.70 & 0 & 17.70 & 11.18 & 0 & 358 & -  & 70.3 \\
\midrule
Bi-Real~\cite{liu2018birealnet} & 1.42 & 16.76 & 1.69 & 0.20 & 10.99 & 17 & \xmark & 56.4 \\
Real-to-Binary~\cite{martinez2020realtobinary} & 1.55 & 16.25 & 1.82 & - & - & - & \checkmark & 65.4 \\
ReActNet~\cite{liu2020reactnet} & 1.44 & 16.76 & 1.70 & 0.21 & 10.99 & 18 & \xmark & 60.2 \\
ReActNet~\cite{liu2020reactnet} & 1.44 & 16.76 & 1.70 & 0.21 & 10.99 & 18 & \checkmark & 65.9 \\
ReCU~\cite{Xu_2021_recu} & 1.44 & 16.76 & 1.70 & 0.20 & 10.99 & 17 & \xmark & 61.0 \\
ReCU~\cite{Xu_2021_recu} & 1.44 & 16.76 & 1.70 & 0.20 & 10.99 & 17 & \checkmark & 66.4 \\
BONN~\cite{zhao2022bonn} & 1.44 & 16.76 & 1.70 & 0.20 & 10.99 & 17 & \xmark & 59.3 \\
BONN~\cite{zhao2022bonn} & 1.44 & 16.76 & 1.70 & 0.20 & 10.99 & 17 & \checkmark & 66.2 \\
\textcolor{black}{BCDNet-A~\cite{xing2022bcdnet}} & \textcolor{black}{0.32} & \textcolor{black}{48.2} & \textcolor{black}{1.08} & \textcolor{black}{-} & \textcolor{black}{-} & \textcolor{black}{-} & \textcolor{black}{\checkmark} & \textcolor{black}{66.9} \\
\textcolor{black}{BCDNet-B~\cite{xing2022bcdnet}} & \textcolor{black}{0.34} & \textcolor{black}{48.2} & \textcolor{black}{1.09} & \textcolor{black}{-} & \textcolor{black}{-} & \textcolor{black}{-} & \textcolor{black}{\checkmark} & \textcolor{black}{67.9}\\
\midrule
\emph{Bi-PiDiNet} & 0.25 & 42.74 & 0.92 & 0.04 & 19.23 & 20 & \xmark & 62.8 \\
\emph{Bi-PiDiNet} & 0.25 & 42.74 & 0.92 & 0.04 & 19.23 & 20 & \checkmark & 66.8 \\

\bottomrule
\end{tabular}
}
\end{center}
\end{table*}

\vspace{0.3em}
\noindent \textcolor{black}{\textbf{Model robustness.} \quad
Similar to the analysis of robustness on edge detection,
we also evaluated the robustness of our Bi-PiDiNet on object recognition.
Specifically, the degraded CIFAR-100 dataset (\emph{i.e.}, CIFAR100-C~\cite{hendrycks2019imagenetc}) was used for evaluation, which involves diverse degradation types like noises, corruptions, and perturbations. From \cref{Tab:noise_cifar}, it can be seen that regardless of the degradation type, the performance of Bi-PiDiNet consistently produces better or on-par performance as compared to the baseline.
This further validates the superior robustness of our Bi-PDC against the vanilla binary convolution.
}

\subsubsection{Comparison with the State-of-the-art Methods}

\noindent \textbf{On ImageNet dataset.} \quad
During training, the input images were augmented with random cropping (to size $224 \times 224$) and random horizontal flipping. The learning rate was set to 0.001, which was reduced by 0.1 at epoch 45 and 55. The batch size was set to 256. Note that, tricks like multi-stage training~\cite{martinez2020realtobinary,liu2020reactnet} and knowledge distillation~\cite{hinton2015distilling}) were not adopted for fair comparison with previous works.

\begin{table*}[t!]
\caption{Testing accuracies and AUC values on facial datasets and the associated network complexities.}
\centering
\setlength{\tabcolsep}{0.01\linewidth}
\resizebox*{\linewidth}{!}{
\begin{tabular}{lccccccccccc}
\toprule
Model & W/A & \begin{tabular}{@{}c@{}}OPs\\($\times 10^8$)\end{tabular} & \begin{tabular}{@{}c@{}}Memory\\(M bits)\end{tabular} & \multicolumn{2}{c}{LFW} & \multicolumn{2}{c}{CALFW} & \multicolumn{2}{c}{CPLFW} & \multicolumn{2}{c}{YTF}\\
\cmidrule(r){5-12}
& & & & ACC (\%) & AUC (\%) & ACC (\%) & AUC (\%) & ACC (\%) & AUC (\%) & ACC (\%) & AUC (\%) \\
\midrule
Full-precision & 32/32 & 17.40 & 373 & 99.22 & 99.89 & 90.78 & 95.68 & 76.18 & 81.94 & 93.20 & 97.80 \\
\midrule
Bi-Real~\cite{liu2018birealnet} & 1/1 & 0.52 & 18 & 96.80 & 99.49 & 81.88 & 89.09 & 67.28 & 71.72 & 90.40 & 96.64 \\
ReActNet~\cite{liu2020reactnet} & 1/1 & 0.52 & 18 & 97.75 & 99.71 & 85.80 & 92.18 & 69.73 & 76.08 & 92.26 & 97.35 \\
\midrule
\emph{Bi-PiDiNet} & 1/1 & 0.52 & 18 & \textbf{98.33} & \textbf{99.84} & \textbf{87.03} & \textbf{93.53} & \textbf{73.47} & \textbf{79.26} & \textbf{92.88} & \textbf{97.87} \\
\bottomrule
\end{tabular}
}
\label{tab:facial}
\end{table*}

We compared our method with thirteen state-of-the-art methods, including ABC-Net~\cite{lin2017abcnet}, XNOR-Net~\cite{rastegari2016xnor}, DoReFa-Net~\cite{zhou2016dorefa}, Bi-Real Net~\cite{liu2018birealnet}, XNOR-Net++~\cite{bulat2019xnorpp}, IR-Net~\cite{qin2020forward}, Han~\etal~\cite{han2020noisysup}, RBNN~\cite{Mingbao2020rbnn}, ReActNet~\cite{liu2020reactnet}, FDA-BNN~\cite{xu2021frequencydomain}, Bi-half~\cite{li2022equalbits}, ReCU~\cite{Xu_2021_recu}, BONN~\cite{zhao2022bonn}. For a fair comparison, we reported the results achieved by different methods with the same training setting as ours (\ie without additional techniques like multi-stage training and knowledge distillation). For ReActNet~\cite{liu2020reactnet}, since its results under this setting were not reported in the original paper, we used our re-implementation for comparison. Quantitative results are shown in \cref{tab:comparison-imagenet}.

It can be observed that our Bi-PiDiNet outperforms all previous methods with notable accuracy improvement. For example, as compared to ReCU, our Bi-PiDiNet produces Top-1 accuracy improvements of 1.8\%/1.2\% for ResNet-18/ResNet-34. This is because our Bi-PDC enhances the representation capacity of our network by capturing high-order information  to produce higher performance. 

The comparison of computational complexity is elaborated in \cref{tab:resnet18-imagenet}. As we can see, the proposed Bi-PiDiNet achieves higher accuracy with nearly halved computational cost as compared to previous approaches. By using additional training techniques, namely, multi-stage training and knowledge distillation, Bi-PiDiNet achieves the best Top-1 accuracy of 66.8\%. 

\textcolor{black}{
To complete our experiments, we additionally replaced Bi-PDC with its full-precision version (\ie PDC) and evaluated the models on ImageNet (all models use the original ResNet-18/34 architectures). The results are shown in \cref{tab:comparison-imagenet}. We can still observe an accuracy gain from the vanilla convolution based model to the PDC equipped version (\eg from 73.5\% to 73.9\% of Top-1 accuracy on ResNet-34).
}

\vspace{0.3em}
\noindent \textbf{On facial datasets.} \quad
As LBP descriptors have been demonstrated to produce promising performances on the facial recognition task~\cite{face-lbp1,ahonen2004face-lbp2,su2019bird}, we further conducted experiments to investigate how the integration of LBP to binary convolutions can enhance the accuracy of BCNNs on this task. 
Specifically, the 20-layer CNN architecture used in~\cite{liu2017sphereface} was employed as the baseline to construct our Bi-PiDiNet, and A-SoftMax loss~\cite{liu2017sphereface} with the angular margin $m=4$ was adopted for training. The initial learning rate was 0.001, which was decayed by 0.1 at epoch 12 and 17. Batch size was set to 128. The models were trained on CASIA-WebFace and evaluated on the other four facial datasets. Two evaluation metrics were applied, including the accuracy under the \emph{unrestricted with labeled outside data protocol}, and the value of AUC. For fair comparison with previous methods, we apply the same architecture and training setting for Bi-Real Net~\cite{liu2018birealnet} and ReActNet~\cite{liu2020reactnet}. The quantitative results are presented in \cref{tab:facial}.

As we can see, our Bi-PiDiNet produces consistent performance gains on 4 benchmark datasets.
For example, compared with ReActNet, our method obtains 0.58\%, 1.23\%, 3.74\%, and 0.62\% accuracy gain on the LFW, CALFW, CPLFW, and YTF datasets, respectively. The performance improvements are more significant in challenging datasets like CALFW and CPLFW where the faces have large variations in ages and facial poses. This further demonstrates that capturing informative high-order facial details using our Bi-PDC are beneficial to the performance.

\begin{figure}[t!]
\captionsetup{labelfont={color=black},font={color=black}}
    \centering
    \includegraphics[width=\linewidth]{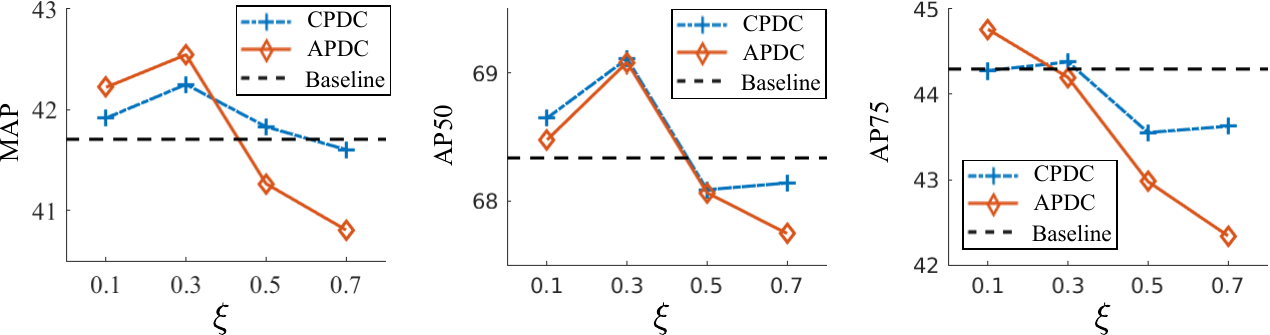}
    \caption{
    Evaluation results on VOC 2007 test set for object detection, where GhostNet was selected as the baseline. The vanilla convolutions in the baseline were gradually replaced with the proposed PDC for comparison.
    }
    \label{fig:od_ap}
\end{figure}

{\color{black}
\subsection{Exploration on object detection}
\label{sec:experiments-objectdetection}

To further explore the generalization of our PDC, we conducted our experiments on the object detection task.
Precisely, we chose GhostNet~\cite{han2020ghostnet} as our baseline and replaced the $3\times 3$ convolutions in the Ghost Modules with our APDC or CPDC (again, we set $\xi$ to control the percentage of convolutions replaced by PDC). All the models were trained from scratch using PASCAL VOC 2007 and 2012 trainval sets with 30K iterations, and tested on VOC 2007 test set. GhostNet is already a lightweight model for object recognition and detection, making it more challenging for PDC to bring further improvement. However, it can be seen from \cref{fig:od_ap} that both CPDC and APDC can strengthen the baseline model with additional performance gain with $\xi=0.3$, proving the effectiveness of our PDC on other tasks like object detection.
}

\section{Conclusion}
\label{sec:conclusion}
This paper proposes two types of convolution named pixel difference convolution (PDC) and binary  pixel difference convolution (Bi-PDC) to enhance the representation capacity of CNNs.
Our (Bi-)PDC incorporates the capability of LBP to capture high-order information complementary to vanilla convolution. Besides, (Bi-)PDC is fully differentiable, computationally efficient, and can be seamlessly embedded into CNN architectures. Based on PDC and Bi-PDC, we designed two lightweight networks named Pixel Difference Network (PiDiNet) and Binary Pixel Difference Network (Bi-PiDiNet) respectively for edge detection and object recognition tasks. Experiments on a wide range of benchmark datasets (BSDS500, ImageNet, LFW, YTF, \etc) show that (Bi-)PDC can facilitate our networks to achieve better trade-off between accuracy and efficiency compared with previous state-of-the-art counterparts.

\vspace{0.3em}
\noindent \textbf{Future work.} \quad
We preserve great space for further exploration on PDC and Bi-PDC.
From the microstructure side, different pattern probing strategies can be investigated to generate (Bi-)PDC instances for certain tasks at hand. From the macrostructure side, a network can be potentially enhanced by optimally configuring multiple (Bi-)PDC instances. With the capability of capturing high-order information, we believe the proposed (Bi-)PDC can benefit more semantically low- and high-level computer vision tasks like salient object detection, facial behavior analysis, object detection, \etc

\section*{Acknowledgment}
This work was supported in part by the National Key Research
and Development Program of China under Grant 2021YFB3100800, in part by
the Academy of Finland under Grant 331883, in part by the Infotech Project
FRAGES, in part by the National Natural Science Foundation of China under
Grants 62376283, 61872379, and 62022091, and in part by the CSC IT Center
for Science, Finland for computational resources.

\footnotesize
\bibliographystyle{IEEEtran}
\bibliography{IEEEabrv,myreference}

%\vspace{-1.5 cm}

\begin{IEEEbiography}[{\includegraphics[width=1in,height=1.25in,clip,keepaspectratio]{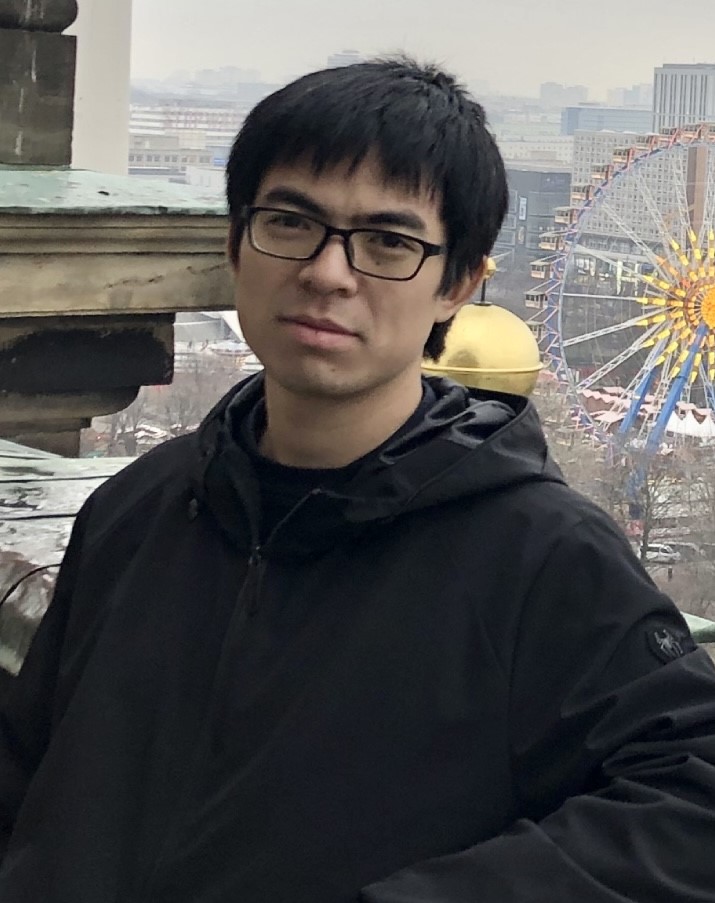}}]{Zhuo Su}
 received the B.E. and M.S. degrees
in School of Automation Science and Electrical Engineering from Beihang University (BUAA), Beijing, China, in 2015 and 2018, respectively. He is
currently pursuing the Ph.D. degree in Computer Science from the Center for Machine Vision and Signal Analysis, University of Oulu, Finland. His research interests include deep learning and machine learning. He is now focusing on network compression and efficient network design for computer vision.
\end{IEEEbiography}

\vspace{-1.0 cm}

\begin{IEEEbiography}[{\includegraphics[width=1in,height=1.25in,clip,keepaspectratio]{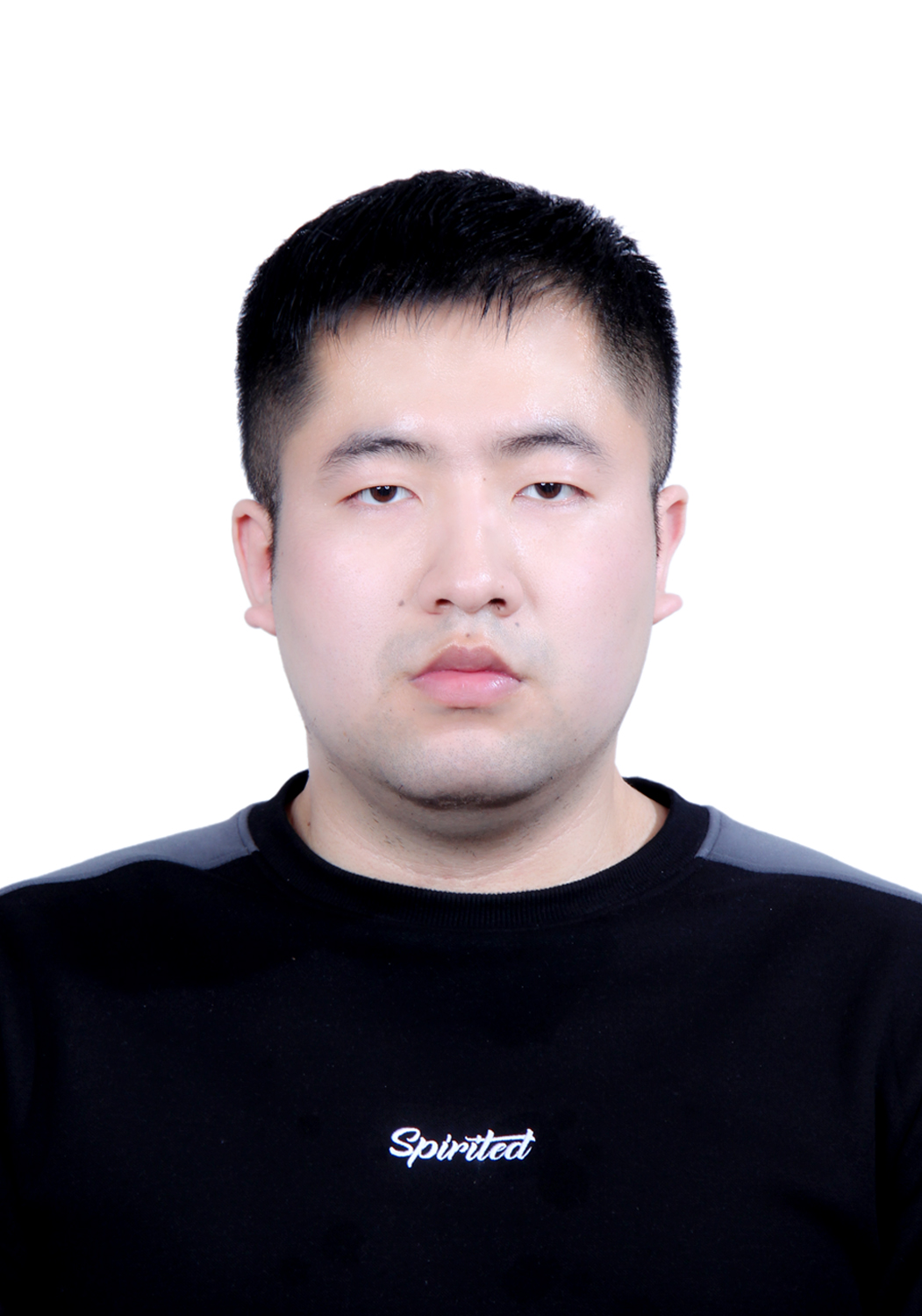}}]{Jiehua Zhang}
received the university B.E. degree in detection, guidance and control technology from University of Electronic Science and Technology of China (UESTC), Chengdu, China, and M.S. in Aeronautical and Astronautical Science and Technology from National University of Defense Technology (NUDT), Changsha, China, in 2018 and 2020, respectively. He is currently pursuing the Ph.D. degree in computer science from University of Oulu, Finland. His research interests include deep learning, computer vision, and efficient network design.
\end{IEEEbiography}

\vspace{-1.0 cm}

\begin{IEEEbiography}[{\includegraphics[width=1in,height=1.25in,clip]{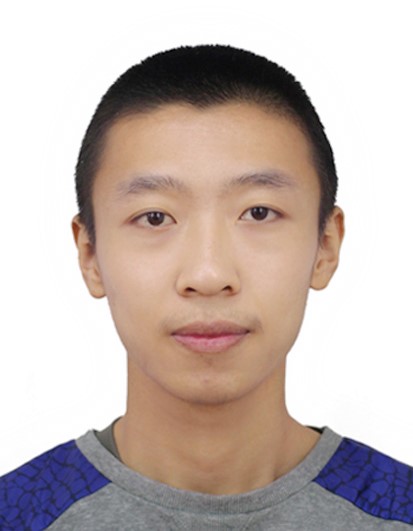}}] {Longguang Wang} received the B.E. degree in Electrical Engineering from Shandong University (SDU), Jinan, China, in 2015, and the Ph.D. degree in Information and Communication Engineering from National University of Defense Technology (NUDT), Changsha, China, in 2022. His current research interests include low-level vision and 3D vision.
\end{IEEEbiography}
	
\vspace{-1.0 cm}

\begin{IEEEbiography}[{\includegraphics[width=1in,height=1.25in,clip,keepaspectratio]{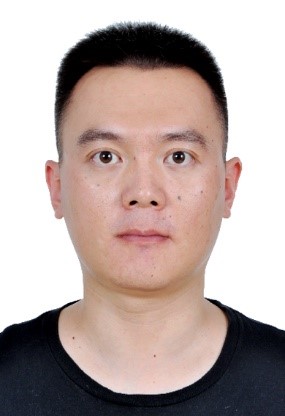}}]{Hua Zhang} is an associate professor with the Institute of Information Engineering, Chinese Academy of Sciences. He received the Ph.D. degrees in computer science from the School of Computer Science and Technology, Tianjin University, Tianjin, China in 2015. His research interests include computer vision, multimedia, and machine learning.
\end{IEEEbiography}

\vspace{-1 cm}

\begin{IEEEbiography}[{\includegraphics[width=1in,height=1.25in,clip,keepaspectratio]{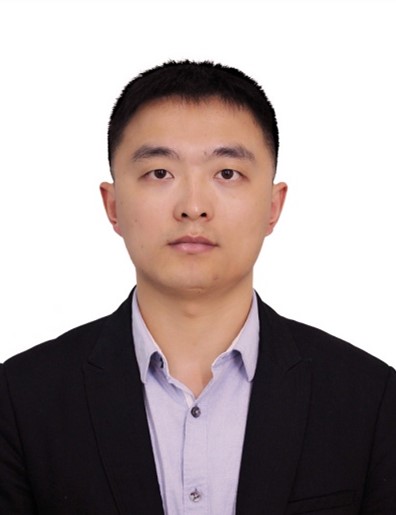}}]{Zhen Liu} received the Ph.D. degree in Information and Communication Engineering from National University of Defense Technology (NUDT), in 2013. He is currently a professor with the College of Electronic Science and Technology, NUDT. He has been awarded the Excellent Young Scientists Fund on his project titled “Intelligent Countermeasure for Radar Target Recognition” in 2020. His current research interests include radar signal processing, radar electronic countermeasure, compressed sensing, and machine learning.
\end{IEEEbiography}

\vspace{-1 cm}

\begin{IEEEbiography}[{\includegraphics[width=1in,height=1.25in,clip,keepaspectratio]{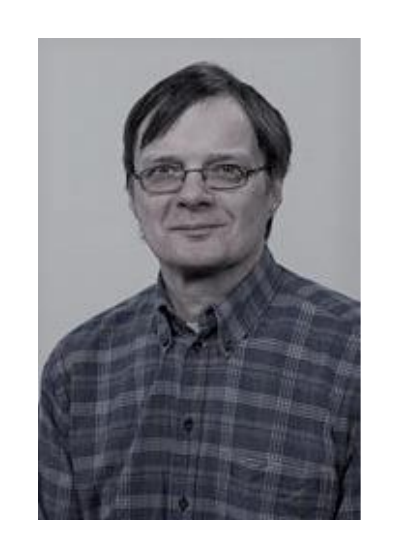}}]{Matti Pietik\"{a}inen} received the doctor of science degree in technology from the University of Oulu, Finland. He is an emeritus professor with the Center for Machine Vision and Signal Analysis, University of Oulu. From 1980 to 1981 and from 1984 to 1985, he visited the Computer Vision Laboratory, University of Maryland. He has made fundamental contributions, \eg, to Local Binary Pattern (LBP) methodology, texture based image and video analysis, and facial image analysis. He has authored more than 350 refereed papers in international journals, books, and conferences. His papers have about 80,000 citations citations in Google Scholar (hindex 98). In 2014, his research on LBP-based face description was awarded the Koenderink Prize for fundamental contributions in computer vision. He was the recipient of the IAPR King-Sun Fu Prize 2018 for fundamental contributions to texture analysis and facial image analysis. He is a fellow of the IEEE for contributions to texture and facial image analysis for machine vision.
\end{IEEEbiography}

\vspace{-1 cm} 

%% author
\begin{IEEEbiography}[{\includegraphics[width=1in,height=1.25in,clip,keepaspectratio]{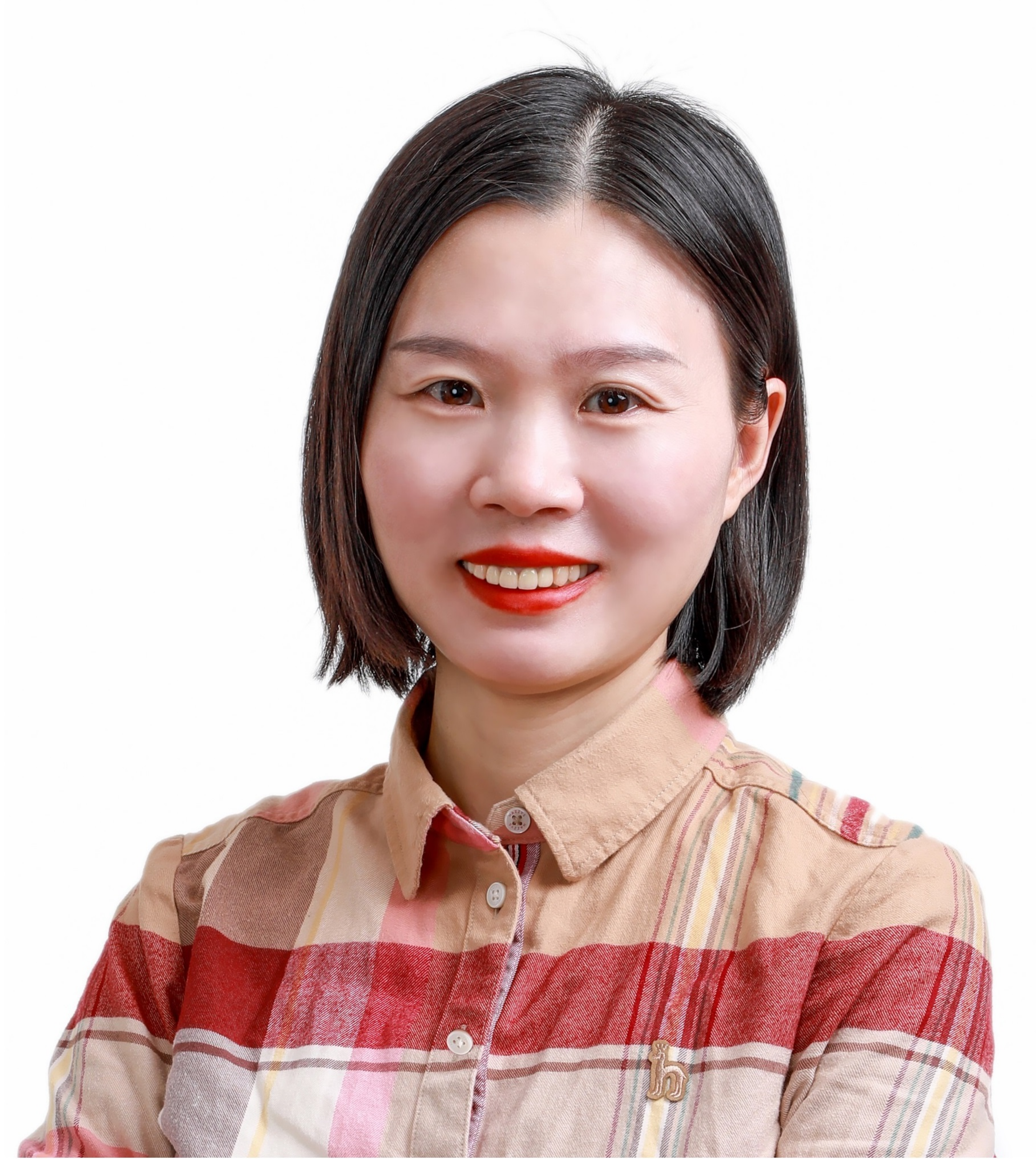}}]{Li Liu} 
received her Ph.D. degree in information and communication engineering from the National University of Defense Technology (NUDT), China, in 2012.She is now a Full Professor with the College of Electronic Science and Technology at NUDT. During her PhD study, she spent more than two years as a Visiting Student at the University of Waterloo, Canada, from 2008 to 2010. From 2015 to 2016, she spent ten months visiting the Multimedia Laboratory at the Chinese University of Hong Kong. From 2016.12 to 2018.11, she worked as a senior researcher at the Machine Vision Group at the University of Oulu, Finland. She was a cochair of nine International Workshops at CVPR, ICCV, and ECCV. She served as the Leading Guest Editor for special issues in \emph{IEEE Transactions on Pattern Analysis and Machine Intelligence (IEEE TPAMI)} and \emph{International Journal of Computer Vision}. She is serving as the Leading Guest Editor for IEEE TPAMI special issue on ``Learning with Fewer Labels in Computer Vision''. Her current research interests include computer vision, pattern recognition and machine learning. Her papers have currently over 10500 citations according to Google Scholar. She currently serves as Associate Editor for \emph{IEEE Transactions on Geoscience and Remote Sensing (IEEE TGRS)}, \emph{IEEE Transactions on Circuits and Systems for Video Technology (IEEE TCSVT)}, and Pattern Recognition.
\end{IEEEbiography}

% that's all folks
\end{document}